\documentclass{article}

\PassOptionsToPackage{numbers,sort&compress}{natbib}
\usepackage[preprint]{neurips_2026}

\usepackage[utf8]{inputenc}
\usepackage[T1]{fontenc}
\usepackage[hidelinks]{hyperref}
\usepackage{booktabs}
\usepackage{amsmath}
\usepackage{amssymb}
\usepackage{amsthm}
\usepackage{microtype}
\usepackage{xcolor}
\usepackage{graphicx}
\usepackage{array}
\usepackage{adjustbox}
\usepackage{placeins}

\title{DART: Semantic Recoverability for Structured Tool Agents}

\author{
  Ke Yang$^{1}$, Panpan Li$^{2}$, Zonghan Wu$^{3}$, Kejin Xu$^{1}$, Huaxi Huang$^{4}$, Xiaoshui Huang$^{5}$\thanks{Corresponding author. Code: \url{https://github.com/KeoYang/DART}.} \\
  $^{1}$MOS Intelligent Connectivity Technology Co. Ltd. \\
  $^{2}$Sichuan Vocational College of Post and Telecom \\
  $^{3}$East China Normal University \\
  $^{4}$Shanghai Artificial Intelligence Laboratory \\
  $^{5}$Shanghai Jiao Tong University
}

\newtheorem{theorem}{Theorem}
\newtheorem{corollary}{Corollary}
\newtheorem{lemma}{Lemma}

\newcommand{\dart}{\textsc{DART}}
\newcommand{\compfrozen}{\textsc{Comp-Frozen}}
\newcommand{\retryonly}{\textsc{Retry-Only}}
\newcommand{\registryonly}{Registry-Only Sidecar}
\newcommand{\registryonlyartifact}{\texttt{registry\_only\_v1}}
\newcommand{\inlinecounterfactualartifact}{\texttt{inline\_snapshot\_manager\_v1}}

\begin{document}

\maketitle

\begin{abstract}

When a structured tool agent fails mid-execution, the runtime faces a dilemma: replaying the entire task is safe but wasteful, while restoring from a local checkpoint is efficient but can leave committed downstream work tied to an upstream history that no longer exists. This tension is acute in commitment-sensitive settings, where rollback targets a single failed instance yet downstream consumers have already acted on its output. Existing recovery approaches provide mechanical rollback but no criterion for whether a local restore remains semantically valid after downstream commitment. We formalize this gap as semantic recoverability and address it in \dart{}, a modular runtime that localizes the failed instance, certifies semantically recoverable boundaries of that instance, aligns checkpoints to those boundaries, and selects an admissible restore point that preserves committed downstream work under dependency and effect constraints—or blocks otherwise. Across three LLM-driven domains and external validation on a LangGraph-based substrate, \dart{} correctly recovers all evaluated commitment-sensitive cases where baseline local recovery fails, and a five-domain safety audit finds no unsafe admitted rollbacks. These results show that controller legality does not imply semantic validity, and that sound local recovery requires an explicit admissibility check.

\end{abstract}

\section{Introduction}
Large language models are increasingly deployed as tool-using agents in production settings such as workflow assistants, scheduling and booking systems, and multi-stage orchestration pipelines.
Failures during execution are inevitable, and the ability to recover without replaying an entire task from scratch is operationally critical.
A broad class of such deployments can be characterized as structured tool agents: tool-using agents whose execution is organized by explicit control flow, observable action boundaries, and persisted traces.
Their explicit structure makes partial-progress reuse feasible, yet existing recovery mechanisms do not check whether a local restore is semantically correct, particularly when downstream work has already been committed.

Current recovery approaches for such systems fall into three broad patterns.
The first assumes that the recoverable object is known in advance: classical workflow-exception and runtime-repair methods define exception scopes, compensation handlers, or service-level regions at design time, so what to recover is never a runtime question \cite{workflowexceptions1999,workflowexceptions2000,guidedrecovery2010,monitoringrecovery2010,smartmonitors2004,automaticworkarounds2010,selfhealingbpel2007}.
The second assumes that the rollback boundary is known in advance: distributed-snapshot and transaction-oriented protocols fix the scope at the process, transaction, or checkpoint level and enforce consistency within it, so how far to roll back is never a runtime question either
\cite{chandy1985snapshot,elnozahy2002rollback,haerder1983recovery,haerder1987nested,garcia1987sagas,featonby2021idempotent}.
The third provides the mechanism but not the criterion: modern agent runtimes such as LangGraph, resume and retry primitives that make local restore mechanically possible, yet they offer no way to determine whether a restored execution is still semantically valid once downstream work has already been committed \citep{langgraph2026persistence,langgraph2026interrupts,langsmith2026rollback,awsstepfunctions2026retry,ray2026fault}.

Across all three lines, recovery faces a fundamental dilemma. Whole-task rerun is always safe but wasteful, because it replays an arbitrary amount of already-completed work. Local restore is efficient but can be semantically invalid when committed downstream consumers remain in place. The root cause is that controller legality, i.e., the runtime's ability to mechanically restore a prior state, does not imply semantic validity: the restored execution may no longer correspond to any valid upstream history. This mismatch manifests in three concrete failure modes: (i) the runtime targets the wrong failed instance, (ii) local rollback invalidates committed downstream work, or (iii) rollback crosses an irreversible effect boundary.

\begin{figure}[t!]
  \centering
  \includegraphics[width=0.85\linewidth]{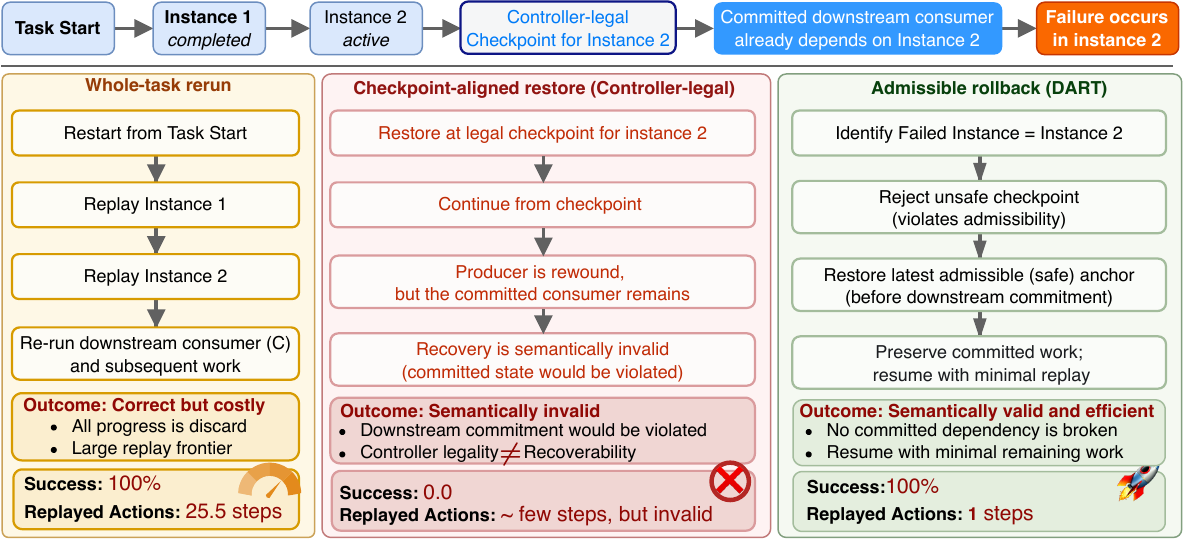}
  \caption{Commitment-sensitive recovery regime. Whole-task rerun is correct but expensive because it replays an unrelated completed prefix. Checkpoint-aligned restore can remain controller-legal yet become semantically invalid when committed downstream consumers remain in place. }
  \label{fig:motivation}
\end{figure}

Consider a scheduling assistant with two subtasks: the first queries three participants' calendars and proposes candidate meeting times, and the second picks one of those times and sends calendar invitations to all participants. Suppose the first subtask completes, the second sends the invitations, and then the first is found to have failed (e.g., a stale calendar cache produced an invalid candidate list). The runtime can roll back the first subtask and retry it. This rollback is controller-legal, but the invitations from the second subtask are already committed and now refer to a time slot that may not appear in any valid retry. Figure~\ref{fig:motivation} illustrates this pattern.
This example exposes a broader limitation: once rollback is localized to a failed semantic unit while committed downstream work remains in place, controller legality alone does not ensure correctness.

We study this gap as a question of semantic recoverability: when a failed instance is rolled back locally, under what conditions is the restore point not only controller-legal but also semantically valid?
Our key idea is to certify, before any local rollback is attempted, that the boundary of the failed instance is semantically closure, meaning that no committed downstream work depends on the specific output being rolled back.
Building on this idea, we introduce \dart{} (Deterministic Agent Runtime with Transition Guards), a modular recovery runtime organized around four decision steps: failed-instance localization, recoverable-boundary certification, instance-aligned checkpointing, and admissible rollback selection.
When all four steps succeed, \dart{} restores the latest admissible local checkpoint. When any step fails, it conservatively blocks the local rollback and falls back to whole-task rerun.
Although we instantiate \dart{} with explicit finite-state-machine (FSM) agents for observability, the underlying problem arises more broadly in any structured runtime where rollback is localized to a failed instance while committed downstream work is preserved.

We evaluate \dart{} across three LLM-driven domains. In the evaluated commitment-sensitive cases, \dart{} recovers every scenario correctly, whereas entry-only restore fails and whole-task rerun incurs substantially larger replay. We reproduce the same contrast on a LangGraph-based runtime: its built-in checkpoint restore fails in the decisive commitment-sensitive case where \dart{} succeeds. Outside commitment-sensitive settings, \dart{} remains competitive with existing approaches, and a systematic safety audit across all evaluated domains confirms that it introduces no unsafe recoveries.

\noindent\textbf{Contributions.}
(1) We formalize semantic recoverability: when is a local rollback not only mechanically possible but also semantically valid? We show that, in commitment-sensitive settings, the two notions diverge and controller legality alone does not prevent invalid recoveries.
(2) To close this gap, we define recoverable boundaries via four conditions (decidability, closure, separability, and controllability) that a rollback target must satisfy, and organize them into a four-step runtime procedure realized in \dart{}.
(3) We prove that the resulting blocking behavior is not merely a conservative design choice but a necessity: any runtime that admits all controller-legal local rollbacks will produce semantically invalid executions in commitment-sensitive settings. Empirical validation on LangGraph confirms this result.

\section{Related Work}
Prior work relevant to structured tool-agent recovery typically fixes the recovery unit, fixes the rollback scope, or provides persistence without an explicit semantic admissibility criterion.

\noindent\textbf{Recovery Units Assumed in Advance.}
Existing work in this area typically assumes that the recoverable unit is already known. Classical workflow systems study exception handling and exception scopes in long-running processes \citep{workflowexceptions1999,workflowexceptions2000}. Runtime-repair work adds guided, monitor-driven, and workaround-based recovery \citep{guidedrecovery2010,monitoringrecovery2010,smartmonitors2004,automaticworkarounds2010}, while self-healing process repair extends this line to adaptive workflow correction \citep{selfhealingbpel2007}. Across these lines, the recovery object is typically authored in advance as an activity, exception scope, or service-level region. Our setting differs in that the runtime must first identify a unique failed semantic instance before any local recovery decision can be made.

\noindent\textbf{Rollback Scope Assumed in Advance.}
Complementary work fixes the rollback boundary in advance and then reasons about consistency within that scope. Distributed snapshots and rollback-recovery protocols restore execution state under a predefined scope \citep{chandy1985snapshot,elnozahy2002rollback}. Transaction-oriented recovery and nested transactions formalize atomicity and scoped rollback \citep{haerder1983recovery,haerder1987nested}, while sagas and idempotent-effect patterns address long-running and externally visible actions \citep{garcia1987sagas,featonby2021idempotent}. These mechanisms are essential for consistency, but they usually assume that rollback scope has already been fixed at the process, transaction, or checkpoint level. By contrast, our question is whether a controller-legal local restore remains semantically admissible under the current dependency and effect context.

\noindent\textbf{Persistence Without Semantic Admissibility.}
More recent runtime systems expose persistence and retry primitives, but still stop short of defining when a local restore is semantically admissible. Modern graph runtimes such as LangGraph expose persistence, interrupts, and rollback-oriented execution primitives \citep{langgraph2026persistence,langgraph2026interrupts,langsmith2026rollback}, and Step Functions and Ray similarly provide retry and fault-tolerance mechanisms \citep{awsstepfunctions2026retry,ray2026fault}. Explicit control structure also provides a more inspectable execution substrate for tool agents \citep{zhang2026evofsm}, while adjacent work studies agent failure diagnosis and debugging \citep{barke2026agentrx,zhu2025agentdebug,in2026rethinking}, self-correction under tool failures \citep{vuddanti2025paladin}, and resilient multi-agent execution \citep{chang2025alas,huang2025resilience}. SagaLLM adds transaction guarantees for multi-agent planning \citep{chang2025sagallm}, and diagnosability-oriented work sharpens fault detection and localization \citep{cassandras2021ides,sampath1995diagnosability}. Together, these works make structured recovery increasingly practical, but they still largely treat correctness as implicit once execution can be retried or restored. DART isolates the missing semantic layer: when a controller-legal local restore is semantically admissible under downstream commitments and effect boundaries.

Taken together, prior work leaves open how to identify the recoverable unit, when a controller-legal checkpoint is also a semantically valid recovery boundary, and whether persistence primitives suffice without dependency- and effect-aware admission.

\section{Problem Setting and Scope}

We fix the execution model, observable failure signal, and subtask-instance recovery unit assumed throughout the paper, using explicit FSMs as a transparent canonical instantiation of the broader class of explicit-control runtimes.

\subsection{Basic Execution Model}

\paragraph{Definition 1 (FSM-governed tool agent).}
An agent in our scope is a tuple
\begin{equation}
\label{eq:agent-model}
\mathcal{G} = (S, A, \delta, M, H)
\end{equation}
where $S$ is a finite state set, $A$ is an action set, $\delta \subseteq S \times A \times S$ is an explicit legal transition relation, $M$ is the runtime memory or context, and $H=(e_1,\ldots,e_T)$ is a recorded step history. Each step
\begin{equation}
\label{eq:step-record}
e_t = (s_t, a_t, s_{t+1}, \Delta m_t)
\end{equation}
exposes at least a current state, an executed action, a successor state, and a memory delta.

Eq.~\eqref{eq:agent-model} fixes the control substrate: agents with explicit, inspectable states and actions. In our experiments these are LLM-based tool agents with explicit FSM control \citep{zhang2026evofsm}. The contribution lies in the recovery layer rather than the model architecture.

\subsection{Observable Failure Events}

\paragraph{Definition 2 (Observable failure event).}
An observable failure event is a tuple
\begin{equation}
\label{eq:failure-event}
f = (t, s, a, \sigma)
\end{equation}
where $t$ is the failed step id, $s$ is the runtime state at failure, $a$ is the failed action, and $\sigma$ is a normalized failure signal exposed at an action boundary and consumable by the recovery runtime.

In our setting, $\sigma$ may correspond to tool exceptions, timeouts, governor denials, missing required inputs, execution-chain exceptions, or explicit contract violations. We therefore do not study silent failure detection, latent semantic error discovery, or general root-cause diagnosis.

\subsection{Subtask Skeletons and Subtask Instances}

\paragraph{Definition 3 (Subtask skeleton).}
A subtask skeleton is a reusable semantic template
\begin{equation}
\label{eq:skeleton-def}
K = (k, S_K^{\mathrm{int}}, S_K^{\mathrm{ent}}, P_K^{\mathrm{com}}, P_K^{\mathrm{exit}}, X_K^{\mathrm{in}}, X_K^{\mathrm{out}}, \pi_K^{\mathrm{eff}})
\end{equation}
where $k$ is the skeleton identifier, $S_K^{\mathrm{int}}$ the internal states, $S_K^{\mathrm{ent}}$ the entry states, $P_K^{\mathrm{com}}$ and $P_K^{\mathrm{exit}}$ the commit/exit predicates, $X_K^{\mathrm{in}}$ and $X_K^{\mathrm{out}}$ the input and output interface keys, and $\pi_K^{\mathrm{eff}}$ the effect policy, i.e., the reviewed rollback policy for effects produced by this skeleton. In the current \dart{}, these fields come from reviewed boundary configurations rather than automatic synthesis. Eq.~\eqref{eq:skeleton-def} is the skeleton-level reviewed recovery contract: its identity and lifecycle fields support failed-instance localization and checkpoint binding, while its predicate, interface, and effect fields support boundary certification and rollback admissibility in Section~4.

\paragraph{Definition 4 (Subtask instance).}
A concrete runtime occurrence of a skeleton is a subtask instance
\begin{equation}
\label{eq:instance-def}
I = (k, \eta, o)
\end{equation}
where $k$ is the skeleton id, $\eta$ the concrete entity id, and $o$ the ordinal for repeated occurrences.

The recoverable unit is the instance in Eq.~\eqref{eq:instance-def}, not the whole task or the skeleton template. Stage-level labels are insufficient once the same stage re-enters.

\subsection{Scope Commitments and Non-Goals}

We study structured tool-agent runtimes in the explicit-control class, instantiated with explicit FSM control, observable action-boundary failures, and subtask-instance recovery; in the current realization, recoverable boundaries come from reviewed boundary configurations rather than automatic discovery. The system does not eliminate semantic review, but concentrates it onto audited boundary, interface, and effect objects. We do not address silent-failure detection, general fault prediction, universal boundary synthesis, or unrestricted rollback under arbitrary irreversible side effects.

\section{Method}

\subsection{Method Overview: Four Recovery Questions}
Checkpoint-based recovery becomes incomplete for failed-instance-local recovery when it does not determine the failed instance, the recoverable boundaries of that instance, and the rollback targets admissible under dependency and effect constraints. \dart{} makes these requirements explicit through four recovery questions: failed-instance localization, recoverable-boundary certification, instance-aligned checkpointing, and admissible rollback selection. Within the scoped setting studied here, these decisions are necessary for preventing semantically invalid recovery.

\begin{figure}[htbp]
  \centering
  \includegraphics[width=0.85\linewidth]{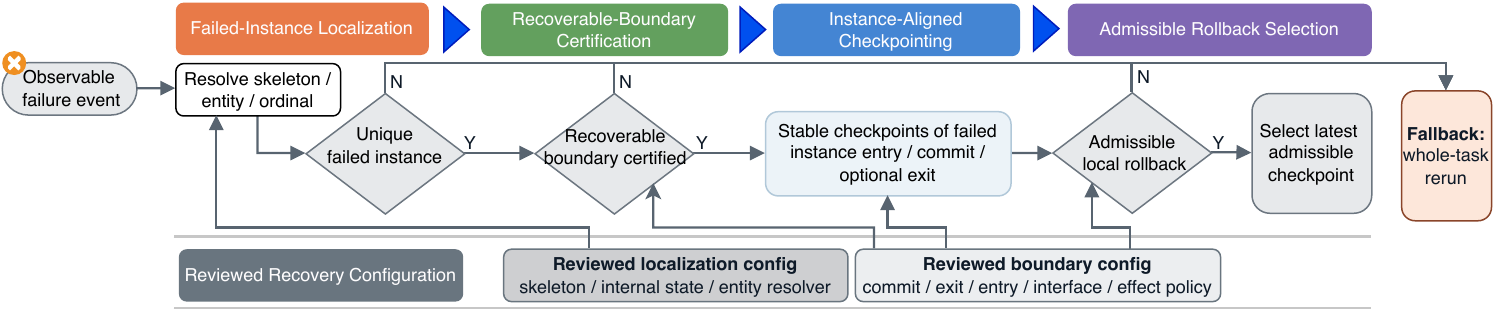}
  \caption{Recovery method overview. After failure, the runtime identifies the failed instance, checks boundary and admissibility conditions, and restores the latest admissible checkpoint; otherwise it falls back to whole-task rerun.}
  \label{fig:method-overview}
\end{figure}

Figure~\ref{fig:method-overview} summarizes this four-layer pipeline. Audits and proof sketches are deferred to Appendix~\ref{app:roadmap} and Appendices~\ref{app:audit} and~\ref{app:proofs}.

\subsection{Failed-Instance Localization: From Observable Failure to Failed Instance}

Building on Definitions~3 and~4, the first step localizes an observable failure to a concrete subtask instance. Given the reviewed skeleton/entity structure, the runtime either identifies a unique failed instance or conservatively abstains, because all later boundary, checkpoint, and rollback judgments are indexed by that instance. Operationally, the runtime resolves the active skeleton, bound entity, and occurrence ordinal from the FSM state, tool arguments, and sidecar registry; if the resulting $(k,\eta,o)$ is not unique, it falls back to conservative whole-task rerun (Table~\ref{tab:appendix-localization-audit}). Localization therefore establishes the instance index consumed by the next layer.

\subsection{Recoverable-Boundary Certification: From Failed Instance to Certified Boundary}

Building on failed-instance localization, the second step certifies which reviewed lifecycle points of that instance are semantically recoverable rather than merely controller-legal. Each reviewed skeleton supplies a minimal recovery contract: an entity resolver, entry states, commit/exit predicates, conservative input/output interface keys, and an effect policy. This step filters reviewed candidates rather than searching over arbitrary legal states or transitions, and it occurs before checkpoint materialization. Informally, a recoverable boundary is a reviewed point from which the failed instance can resume without invalidating the surrounding execution.

\paragraph{Definition 5 (Recoverable boundary).}
Let $b$ be a reviewed commit- or exit-level state or transition associated with subtask instance $I$. We call $b$ a recoverable boundary iff
\begin{equation}
\label{eq:formal-boundary}
\begin{aligned}
\mathrm{Recoverable}(b,I) \iff {} & \mathrm{Decidable}(b,I) \land \mathrm{Closed}(b,I) \\
& \land \mathrm{Separable}(b,I) \land \mathrm{Controllable}(b,I)
\end{aligned}
\end{equation}
Operationally, each conjunct is a concrete test. $\mathrm{Decidable}$ passes iff the candidate still maps to one unique live instance identifier $(k,\eta,o)$. $\mathrm{Closed}$ passes iff the reviewed commit- or exit-level predicate holds and the declared interface handoff is semantically complete. $\mathrm{Separable}$ passes iff restoring from that point keeps replay confined to the failed instance rather than reopening an unrelated task prefix, enforced operationally by binding checkpoints to that instance and restricting restore search to its checkpoint set. $\mathrm{Controllable}$ passes iff the effect policy allows rollback across that frontier. A candidate is certified only if all four tests pass.

This definition separates controller legality from recoverability. A legal lifecycle point may still fail certification because the runtime can no longer tell which instance failed, the current step has not yet reached a semantically complete handoff, replay would reopen work outside that instance, or rollback is disallowed by the effect policy. Operationally, the runtime evaluates only reviewed commit- and exit-level candidates for the failed instance and certifies those that keep the instance identifiable, the handoff closed, replay local, and rollback allowed under the effect policy. For example, the transition from \texttt{WAITING\_POI\_SELECTION} to \texttt{STOP\_READY} is legal but not a certified exit boundary because the current stop instance has not reached a reviewed closed handoff.

\subsection{Instance-Aligned Checkpointing: From Certified Boundary to Stable Recovery Anchor}

Once Section~4.3 certifies recoverable boundaries for the failed instance, the next step is instance-aligned checkpointing. Its objective is to turn those certified boundaries, together with conservative entry anchors, into concrete restore objects indexed by that instance. Given the failed instance and these reviewed lifecycle points, the runtime constructs a recency-ordered checkpoint set attached to that instance. This step introduces no new semantic criterion: it materializes the recoverable structure certified in Section~4.3 rather than re-deciding recoverability.

Operationally, the runtime indexes checkpoint records by instance identity. When an instance becomes active, it records an entry checkpoint; when the instance later satisfies a reviewed commit predicate, it appends a commit checkpoint; and it optionally records an exit checkpoint when a reviewed exit-boundary predicate holds.

\paragraph{Stable checkpoint.}
A stable checkpoint is a concrete restore object attached to a subtask instance and a reviewed lifecycle type. We denote it by
\begin{equation}
\label{eq:checkpoint-def}
c = (I, \tau)
\end{equation}
where $I$ is the subtask instance and $\tau \in \{\texttt{entry}, \texttt{commit}, \texttt{exit}\}$ is the lifecycle type. We write $\mathcal{C}(I)$ for the recency-ordered checkpoint set attached to $I$.
Eq.~\eqref{eq:checkpoint-def} binds each checkpoint to its instance.

Instance-aligned checkpointing matters because later restore search is confined to $\mathcal{C}(I_f)$ for the failed instance $I_f$ rather than collapsing back to an unrelated whole-task prefix. Entry checkpoints provide conservative restart anchors, commit checkpoints preserve stabilized partial progress within the same instance, and exit checkpoints record reviewed completion boundaries for policy-controlled recovery. By default, restore selection considers only entry and commit checkpoints; exit checkpoints are recorded but used only under an explicit policy.

\subsection{Admissible Rollback Selection: From Stable Anchor to Admissible Local Rollback}

Building on the failed instance's stable checkpoint set from Section~4.4, the final step is admissible rollback selection. Its objective is to determine which stable checkpoints of the failed instance remain safe to restore under dependency and effect constraints, and to select the latest such anchor when one exists. The input is the observable failure event together with the failed instance and its stable checkpoints; the output is either an admissible local rollback target or rejection. This step is necessary because restoring from a stable checkpoint may still invalidate committed downstream work or cross an irreversible effect boundary.

\paragraph{Definition 6 (Admissible local recovery).}
Given an observable failure event $f$, a recovery is admissible local recovery iff there exist a subtask instance $I$ and a stable checkpoint $c$ such that
\begin{equation}
\label{eq:safe-recover}
\begin{aligned}
\mathrm{AdmissibleRecover}(f,I,c) \iff {} & \mathrm{Identified}(f,I) \land \mathrm{Stable}(c,I) \\
& \land \mathrm{ScopeOK}(I,c) \land \mathrm{NoCommittedConflict}(I) \\
& \land \mathrm{EffectAllowed}(I,c)
\end{aligned}
\end{equation}
In brief, $\mathrm{Identified}$ fixes the failed instance, $\mathrm{Stable}$ requires a checkpoint attached to that instance, $\mathrm{ScopeOK}$ keeps restoration within instance scope, $\mathrm{NoCommittedConflict}$ preserves committed downstream consumers, and $\mathrm{EffectAllowed}$ respects the reviewed effect policy.

For the identified failed instance $I_f$ and failure event $f$, let
\begin{equation}
\label{eq:admissible-set}
\mathcal{A}(f) = \{ c \in \mathcal{C}(I_f) \mid \mathrm{AdmissibleRecover}(f,I_f,c) \}
\end{equation}
denote the admissible checkpoint set for the failed instance. Eq.~\eqref{eq:admissible-set} makes explicit that admissibility is defined over the failed instance's own stable checkpoints rather than the whole task.

After the failed instance and its checkpoint set are fixed, the runtime realizes this gate through dependency- and effect-aware vetoes over $\mathcal{C}(I_f)$. \emph{NoCommittedConflict} is enforced by a conservative producer--consumer relation over instance-level read/write sets induced by the reviewed interface contract, rejecting rollback once a committed downstream instance has consumed the failed instance's outputs. \emph{EffectAllowed} gates candidates by the frozen effect policy $\pi_K^{\mathrm{eff}}$, blocking rollback across disallowed effect boundaries (Appendix~\ref{app:runtime-realization}). The runtime restores the most recent checkpoint in $\mathcal{C}(I_f)$ that satisfies scope, committed-consumer, and effect-policy checks; if none exists, $\mathcal{A}(f)=\emptyset$ and local rollback is rejected. Committed-consumer blocking is therefore necessary for sound local rollback once committed downstream consumers remain in place.
\begin{equation}
\label{eq:checkpoint-select}
c^\star(f) = \max \mathcal{A}(f)
\end{equation}
where the maximum is taken with respect to checkpoint recency within instance $I_f$. Appendix~\ref{app:proofs} gives the remaining statements for this rule.

\section{Experiments}

\subsection{Experimental Setup}

\paragraph{Domains and Protocol.}
We evaluate \dart{} on three LLM-driven domains: navigation, schedule-form, and diagnosis. Two deterministic domains---ETL pipeline and travel planning\citep{xie2024travelplanner}---are deferred to Appendix~\ref{app:generalization}. Across domains, failures are injected only at controlled observable action boundaries, and reviewed boundary, interface/effect, and audit specifications are frozen before evaluation.

\paragraph{Baselines.}
We compare four recovery strategies under a matched runtime and failure protocol: whole-task rerun (\retryonly{}); Coarse-State-Retry, which restores the latest pre-entry snapshot at a benchmark-defined coarse FSM anchor; Comp-EntryOnly, which restores the failed instance's entry checkpoint; and \compfrozen{}, which restores the latest admissible reviewed checkpoint of the failed instance. These comparisons isolate recovery-policy differences on a shared execution substrate. We report both official headline cases and commitment-sensitive cases, where recovery must preserve committed progress and downstream dependencies.

\paragraph{Metrics and Reporting.}
Primary metrics are success, recovery-observed rate, failure-to-milestone latency, replay actions, upstream replay, and preserved completed instances; the semantic audit additionally reports safe-equivalence and admission/blocking statistics. Main-text latency and replay are medians over successful runs; paired tests, no-failure overhead, and full audit details are deferred to Appendix~\ref{app:stats} and Appendix~\ref{app:audit}. We report commitment-sensitive cases first and headline cases second. Extended LangGraph details are in Appendix~\ref{app:langgraph}.

\subsection{In-Domain Recovery Results}

\paragraph{Commitment-Sensitive Recovery: Entry-Only Fails, Certified Checkpoints Succeed.}
We first examine commitment-sensitive failures, where checkpoint choice changes outcome beyond cost. Across commitment-sensitive cases in navigation, schedule-form, and diagnosis, entry-only recovery fails in two ways: in navigation it is observed yet fails the end-to-end task contract, whereas in schedule-form and diagnosis no local recovery is observed with entry-only. In schedule-form, Coarse-State-Retry is omitted because after durable submit point no fair coarse anchor remains beyond the entry-only family. The decisive result is not that \dart{} improves recovery quality by degree, but that entry-only recovery fails in all evaluated commitment-sensitive core-domain cases, whereas reviewed admissible checkpoints succeed throughout. This indicates that without an admissibility criterion, failed-instance-local rollback leads to systematic failure not isolated errors.

\begin{table*}[!htbp]
\caption{Recovery outcomes in the three core LLM domains. Panel~A shows the commitment-sensitive regime; for schedule-form, Coarse-State-Retry is omitted because after submission no fair coarse anchor remains beyond the entry-only family. Panel~B shows official headline cases. Status distinguishes successful completion, contract failure after attempted recovery, no local recovery observed, and explicit blocking when no admissible checkpoint exists. Latency and replay are medians over successful runs.}
\label{tab:recovery-main}
\centering
\scriptsize
\setlength{\tabcolsep}{4pt}
\begin{adjustbox}{max width=\textwidth}
\begin{tabular}{llllrrrrrr}
\toprule
Panel A & Domain & Method & Success & Replay & Up. replay & Preserved inst. & F$\rightarrow$M (ms) & Recov. obs. & Status \\
\midrule
\textbf{Commit-sensitive} & Navigation & \retryonly{} & 1.00 & 18.0 & 14.0 & 0.0 & 25752.95 & 0.00 & ok \\
\textbf{Commit-sensitive} & Navigation & Coarse-State-Retry & 1.00 & 4.0 & 0.0 & 2.0 & 7013.64 & 1.00 & ok \\
\textbf{Commit-sensitive} & Navigation & Comp-EntryOnly & 0.00 & -- & -- & -- & -- & 1.00 & contract \\
\textbf{Commit-sensitive} & Navigation & \textbf{\compfrozen{}} & \textbf{1.00} & \textbf{1.0} & \textbf{0.0} & \textbf{2.0} & \textbf{2527.10} & \textbf{1.00} & \textbf{ok} \\
\textbf{Commit-sensitive} & Schedule Form & \retryonly{} & 1.00 & 29.0 & 24.0 & 0.0 & 34690.47 & 0.00 & ok \\
\textbf{Commit-sensitive} & Schedule Form & Comp-EntryOnly & 0.00 & -- & -- & -- & -- & 0.00 & no-recov \\
\textbf{Commit-sensitive} & Schedule Form & \textbf{\compfrozen{}} & \textbf{1.00} & \textbf{1.0} & \textbf{0.0} & \textbf{5.0} & \textbf{1141.17} & \textbf{1.00} & \textbf{ok} \\
\textbf{Commit-sensitive} & Diagnosis & \retryonly{} & 1.00 & 16.5 & 10.5 & 0.0 & 21174.59 & 0.00 & ok \\
\textbf{Commit-sensitive} & Diagnosis & Coarse-State-Retry & 1.00 & 5.0 & 0.0 & 2.5 & 5878.80 & 1.00 & ok \\
\textbf{Commit-sensitive} & Diagnosis & Comp-EntryOnly & 0.00 & -- & -- & -- & -- & 0.00 & no-recov \\
\textbf{Commit-sensitive} & Diagnosis & \textbf{\compfrozen{}} & \textbf{1.00} & \textbf{2.0} & \textbf{0.0} & \textbf{2.5} & \textbf{2113.80} & \textbf{1.00} & \textbf{ok} \\
\midrule
Panel B & Domain & Method & Success & Replay & Up. replay & Preserved inst. & F$\rightarrow$M (ms) & Recov. obs. & Status \\
\midrule
\textbf{Official} & Navigation & \retryonly{} & 1.00 & 10.0 & 6.0 & 0.0 & 6920.20 & 0.00 & ok \\
\textbf{Official} & Navigation & Coarse-State-Retry & 1.00 & 5.0 & 0.0 & 1.0 & 4735.49 & 1.00 & ok \\
\textbf{Official} & Navigation & Comp-EntryOnly & 1.00 & 5.0 & 0.0 & 1.0 & 4640.04 & 1.00 & ok \\
\textbf{Official} & Navigation & \textbf{\compfrozen{}} & \textbf{1.00} & \textbf{5.0} & \textbf{0.0} & \textbf{1.0} & \textbf{4841.55} & \textbf{1.00} & \textbf{ok} \\
\textbf{Official} & Schedule Form & \retryonly{} & 1.00 & 21.0 & 15.0 & 0.0 & 18530.74 & 0.00 & ok \\
\textbf{Official} & Schedule Form & Coarse-State-Retry & 1.00 & 4.5 & 0.0 & 3.5 & 4410.20 & 1.00 & ok \\
\textbf{Official} & Schedule Form & Comp-EntryOnly & 1.00 & 4.5 & 0.0 & 3.5 & 4176.20 & 1.00 & ok \\
\textbf{Official} & Schedule Form & \textbf{\compfrozen{}} & \textbf{1.00} & \textbf{4.5} & \textbf{0.0} & \textbf{3.5} & \textbf{4431.35} & \textbf{1.00} & \textbf{ok} \\
\textbf{Official} & Diagnosis & \retryonly{} & 1.00 & 11.0 & 5.0 & 0.0 & 12577.31 & 0.00 & ok \\
\textbf{Official} & Diagnosis & Coarse-State-Retry & 1.00 & 5.0 & 0.0 & 1.0 & 6011.75 & 1.00 & ok \\
\textbf{Official} & Diagnosis & Comp-EntryOnly & 1.00 & 5.0 & 0.0 & 1.0 & 6157.04 & 1.00 & ok \\
\textbf{Official} & Diagnosis & \textbf{\compfrozen{}} & \textbf{1.00} & \textbf{5.0} & \textbf{0.0} & \textbf{1.0} & \textbf{5736.50} & \textbf{1.00} & \textbf{ok} \\
\bottomrule
\end{tabular}
\end{adjustbox}
\end{table*}

To ensure robustness, we repeat the decisive commitment-sensitive row across model families. The pattern persists: entry-only restore fails, \compfrozen{} succeeds with one-step replay, and retry-only requires full upstream replay. Because \emph{Replay} and \emph{Up.\ replay} are frontier-size metrics, they are driven by recovery structure not model family (Appendix Table~\ref{tab:xmodel-decisive}).

\paragraph{Official Headline Recovery: Competitive Beyond Commitment-Sensitive Cases.}
Outside the commitment-sensitive regime, the result is narrower. Panel~B of Table~\ref{tab:recovery-main} shows that on headline cases \compfrozen{} remains competitive, preserves zero upstream replay relative to whole-task rerun, and is mostly at parity with stronger local baselines rather than uniformly dominant. Relative to \retryonly{}, latency remains significantly lower in paired analysis (Appendix~\ref{app:stats}).

\subsection{Cross-Runtime External Validation}
We test if the same commitment-sensitive failure pattern reappears once persistence and resume are available in an external LangGraph-based runtime. Table~\ref{tab:langgraph-main-anchor} reports an aligned three-way comparison across \retryonly{}, LangGraph-SemiReal, and \dart{} on regime-specific intersections. The decisive schedule-form commitment-sensitive row is a counterexample: \retryonly{} succeeds with large replay frontier, LangGraph-SemiReal drops to 0.00, and \dart{} remains admissible, succeeds with one-step replay frontier. This shows that the failure is not a runtime artifact, but a general limitation of checkpoint-aligned recovery in commitment-sensitive local rollback settings.

\begin{table*}[t]
\caption{Cross-runtime external validation on aligned regime-specific intersections. In the decisive schedule-form commitment-sensitive case, LangGraph-based checkpoint-aligned restore fails, whereas \dart{} remains admissible and succeeds with a one-step replay frontier.}
\label{tab:langgraph-main-anchor}
\centering
\scriptsize
\setlength{\tabcolsep}{3pt}
\begin{adjustbox}{max width=\textwidth}
\begin{tabular}{llrrrrrrrrr}
\toprule
Domain & Regime & Retry success & LangGraph success & DART success & Retry replay & LangGraph replay & DART replay & Retry F$\rightarrow$M & LangGraph F$\rightarrow$M & DART F$\rightarrow$M \\
\midrule
Navigation & Entry-aligned & 1.00 & 1.00 & 1.00 & 10.0 & 1.5 & 4.0 & 6376.97 & 588.89 & 3328.69 \\
Navigation & Commit-sensitive & 1.00 & 1.00 & 1.00 & 12.0 & 1.0 & 1.0 & 14596.55 & 1938.46 & 2433.90 \\
Schedule Form & Entry-aligned & 1.00 & 1.00 & 1.00 & 21.5 & 4.0 & 6.5 & 19093.78 & 4969.47 & 6137.15 \\
Schedule Form & Commit-sensitive & 1.00 & 0.00 & 1.00 & 25.5 & 0.0 & 1.0 & 32618.41 & -- & 1109.80 \\
\bottomrule
\end{tabular}
\end{adjustbox}
\end{table*}

The navigation rows and the schedule-form entry-aligned row serve as controls, showing that the gap appears where recovery depends on semantic admissibility beyond checkpoint alignment. Appendix~\ref{app:langgraph} further separates portability from necessity through a transplant-control study and blocking witness.

\subsection{Semantic Audit and Blocking Calibration}

The final main-text question is whether admitted recoveries remain semantically acceptable and conservatively calibrated. Table~\ref{tab:correctness-safety} reports 54 \emph{comparable rows} and 47 \emph{evaluated recovery events}; full denominator details are deferred to Appendix~\ref{app:audit}.

\begin{table*}[!htbp]
\caption{Semantic audit and blocking calibration. Panel~A summarizes the five-domain audit for the three core LLM-driven domains shown here; full breakdowns appear in Appendix~\ref{app:audit}. Panel~B reports blocking calibration. ``--'' indicates not applicable for that audit slice.}
\label{tab:correctness-safety}
\centering
\scriptsize
\setlength{\tabcolsep}{3pt}
\begin{adjustbox}{max width=\textwidth}
\begin{tabular}{llrrrrrll}
\toprule
Panel & Scope & Rows / events & Safe-equiv. & Blocked / unsafe & Semantic & Prefix & Effect & Notes \\
\midrule
\textbf{A} & Overall semantic audit & 54 & 1.00 & -- & 54 & 54 & 31 & 21 rows also admit committed-prefix checks \\
\textbf{A} & Navigation & 12 & 1.00 & -- & 12 & 12 & 0 & safe-equivalent on all comparable rows \\
\textbf{A} & Schedule Form & 11 & 1.00 & -- & 11 & 11 & 0 & safe-equivalent on all comparable rows \\
\textbf{A} & Diagnosis & 10 & 1.00 & -- & 10 & 10 & 10 & repair outcomes remain semantically aligned \\
\midrule
\textbf{B} & Overall calibration & 47 & -- & 12 blocked / 0 unsafe & -- & -- & -- & 35 admitted; 0/12 false-blocked events \\
\textbf{B} & Reason family & 12 blocked & -- & 7 effect / 5 dependency & -- & -- & -- & blocking is structured, not random \\
\textbf{B} & Unsafe admission audit & 35 admitted & -- & 0 unsafe / 0.0 rate & -- & -- & -- & 0/35 unsafe admissions \\
\bottomrule
\end{tabular}
\end{adjustbox}
\end{table*}

All 54 comparable rows are safe-equivalent, and the 47-event calibration yields 35 admitted and 12 blocked events with 0/35 unsafe admissions and 0/12 audited false blocks.

\subsection{Ablation Studies}
We conduct ablation studies aligned with our main claims: instance-aligned checkpointing, recoverable boundary certification, and committed-consumer blocking. Table~\ref{tab:mechanism-ablation} reveals that coarse retry remain overly broad without instance-aligned checkpoints (A), controller-legal points may still fail boundary certification (B), removing committed-consumer blocking permits downstream-invalidating rollbacks (C). Collectively, DART’s gains stem from semantic admissibility, not checkpointing alone.

\begin{table*}[t]
\caption{Necessity ablations for semantic admissibility. Panel~A tests instance-aligned checkpointing, Panel~B tests recoverable-boundary certification, and Panel~C tests committed-consumer blocking. Panel~A uses representative cases; aggregates appear in Table~\ref{tab:recovery-main}.}
\label{tab:mechanism-ablation}
\centering
\scriptsize
\setlength{\tabcolsep}{3pt}
\begin{adjustbox}{max width=\textwidth}
\begin{tabular}{llllllll}
\toprule
Panel & Domain / setting & \textbf{\compfrozen{}} & Coarse / Entry-only & \retryonly{} & Conclusion & Key signal & Note \\
\midrule
\textbf{A} & Navigation / entry-aligned & replay 5, latency 3122 & replay 5, latency 3122 & replay 11, latency 6359 & reviewed commit not yet needed & coarse = frozen & same entry anchor \\
\textbf{A} & Navigation / commit-sensitive & replay 1, latency 872 & replay 4, latency 3751 & replay 14, latency 9432 & commit checkpoint adds real gain & frozen strictly smaller replay & preserved inst. = 2 \\
\textbf{A} & Schedule / commit-sensitive & replay 1, latency 50 & entry-only blocked & replay 26, latency 14247 & commit checkpoint remains necessary & entry-only not admissible & preserved inst. = 5 \\
\textbf{A} & Diagnosis / commit-sensitive & replay 2, latency 40 & replay 5 / entry-only fails & replay 15, latency 6430 & commit checkpoint shrinks frontier & coarse still wider than frozen & preserved inst. = 2 \\
\midrule
\textbf{B} & Navigation wrong edge & forced exit emitted & frozen exit absent & legal edge & wrong boundary is unsafe & unresolved branch marked EXITED & \texttt{WAITING\_POI\_SELECTION$\rightarrow$STOP\_READY} \\
\textbf{B} & Schedule wrong edge & forced exit emitted & frozen exit absent & legal edge & wrong boundary is unsafe & unresolved slot marked EXITED & \texttt{WAITING\_SLOT\_SELECTION$\rightarrow$SLOT\_READY} \\
\midrule
\textbf{C} & Navigation consumer blocking & dropped 1 committed consumer & allowed without guard & blocked under guard & blocking is necessary & scope silently expands & downstream stop invalidated \\
\textbf{C} & Schedule consumer blocking & dropped 2 committed consumers & allowed without guard & blocked under guard & blocking is necessary & finalized schedule invalidated & irreversible consumer present \\
\textbf{C} & Diagnosis consumer blocking & dropped 1 committed consumer & allowed without guard & blocked under guard & blocking is necessary & finalized repair invalidated & irreversible repair consumer \\
\bottomrule
\end{tabular}
\end{adjustbox}
\end{table*}

\section{Discussion and Limitations}
\label{sec:discussion}

Our empirical claims are scoped to observable-failure recovery in structured tool agents under reviewed boundary configurations and the current dependency abstraction. This scope is chosen for auditability and transparency, not because the question is unique to explicit-FSM controllers. Commitment-sensitive failures need not dominate all workloads, but they become structurally unavoidable whenever a persisted runtime combines instance-local rollback with independently committed downstream dependencies or effects. In that setting, separating controller legality from semantic recoverability is a correctness requirement. For deployed agent runtimes, the practical implication is that persistence primitives alone are insufficient in commitment-sensitive settings. Local recovery should be attempted only when semantic admissibility can be justified; otherwise, execution-legal rollback may still be globally inconsistent. \dart{} shows that such checks can be layered on top of existing explicit-control runtimes while preserving local progress when safe.

\paragraph{Broader Impacts.}
Explicit local recovery reduces unnecessary tool invocations, preserve completed progress, and improve long multi-stage workflows. Mis-specified boundaries or overly aggressive rollback could still conceal errors in high-stakes settings, so reviewed boundary, conservative blocking, and explicit effect policies remain safeguards rather than mere optimization choices.

\section{Conclusion}
This paper addresses semantic recoverability in structured tool-agent runtimes under preserved downstream commitments. \dart{} makes this problem explicit through failed-instance localization, recoverable-boundary certification, instance-aligned checkpointing, and admissible rollback selection. Empirically, \dart{} improves recovery correctness on the decisive commitment-sensitive cases where baseline local recovery fails, reduces replay relative to whole-task rerun, and introduces no unsafe admitted rollbacks under external LangGraph validation and a five-domain safety audit. More broadly, the results suggest that persistence primitives are necessary but not sufficient for sound local recovery: structured runtimes that preserve downstream committed work need an explicit admissibility criterion.

\appendix

\section{Appendix Roadmap}
\label{app:roadmap}
The appendix is organized as a compact support map rather than a second narrative. The main text now includes a dedicated Discussion and Limitations section that clarifies the scope of our claims, the main limitations of the current study, and the broader implications of the proposed recovery criterion. The appendix provides the remaining supporting evidence: Appendix~\ref{app:xmodel-decisive} gives cross-model robustness on the decisive commitment-sensitive row; Appendix~\ref{app:stats} strengthens the main recovery tables with paired statistics and overhead diagnostics; Appendix~\ref{app:langgraph} provides \emph{external validation} on LangGraph-based runtimes; and Appendix~\ref{app:audit} supplies the main safety-and-mechanism backbone through the five-domain audit chain, localization audit, and boundary/property evidence. Appendix~\ref{app:generalization}, Appendix~\ref{app:proofs}, and Appendix~\ref{app:repro} provide setup breadth, deferred proof support, and reproducibility details.

\begin{table}[!htbp]
\caption{Appendix roadmap by reviewer concern.}
\label{tab:appendix-roadmap}
\centering
\scriptsize
\setlength{\tabcolsep}{3pt}
\begin{adjustbox}{max width=\linewidth}
\begin{tabular}{>{\raggedright\arraybackslash}p{0.34\linewidth}>{\raggedright\arraybackslash}p{0.34\linewidth}>{\raggedright\arraybackslash}p{0.22\linewidth}}
\toprule
Reviewer concern & Evidence type & Where addressed \\
\midrule
What is the scope of the main claim, and what are the study's limitations? & scope commitments, limitations, and broader implications & Section~\ref{sec:discussion} \\
Does the decisive-row result depend on a particular model family? & cross-model decisive-row robustness table & Appendix~\ref{app:xmodel-decisive} \\
Does the result generalize beyond the three core LLM-driven domains? & five-domain universe; deterministic ETL and travel-planning results & Appendix~\ref{app:generalization} \\
Are the latency and replay gains statistically robust? & paired tests, no-failure overhead, checkpoint-granularity diagnostic & Appendix~\ref{app:stats} \\
Is this more than ordinary graph persistence or resume? & LangGraph external validation, transplant-control transportability check, and blocking witness & Appendix~\ref{app:langgraph} \\
Are local recoveries semantically safe? & semantic audit, blocking calibration, boundary evidence, property-wise necessity & Appendix~\ref{app:audit} \\
Is failed-instance localization actually reliable? & repeat-level alignment, ambiguity benchmark, consequence probes & Appendix~\ref{app:audit} (localization audit) \\
What assumptions support the proof-backed claims? & scoped theorem statements and proof sketches & Appendix~\ref{app:proofs} \\
\bottomrule
\end{tabular}
\end{adjustbox}
\end{table}

\section{Cross-Model Decisive-Row Evidence}
\label{app:xmodel-decisive}

Appendix Table~\ref{tab:xmodel-decisive} reports the full cross-model results on the decisive schedule-form commitment-sensitive row used in the main text.

\begin{table*}[!htbp]
\caption{Cross-model results on the decisive schedule-form commitment-sensitive row
(\texttt{schedule\_live\_final\_render\_failure\_after\_submitted}).
Across all tested model families, entry-only restore fails, frozen local recovery succeeds with one-step replay and zero upstream replay, and retry-only succeeds only by replaying the full upstream prefix. Columns are ordered by priority from left to right: success, replay width, upstream replay, preserved progress, and failure-to-milestone latency.}
\label{tab:xmodel-decisive}
\centering
\scriptsize
\setlength{\tabcolsep}{3pt}
\begin{adjustbox}{max width=\textwidth}
\begin{tabular}{llccccc}
\toprule
Model & Method & Success & Replay & Up. replay & Preserved & F$\rightarrow$M (ms) \\
\midrule
GLM & Comp-EntryOnly & 0.00 & 0 & 0 & 5 slots & 0.87 \\
GLM & \retryonly{} & 1.00 & 26 & 21 & 0 slots & 31937.07 \\
GLM & \textbf{\compfrozen{}} & \textbf{1.00} & \textbf{1} & \textbf{0} & \textbf{5 slots} & \textbf{926.92} \\
\midrule
GPT & Comp-EntryOnly & 0.00 & 0 & 0 & 5 slots & 1.42 \\
GPT & \retryonly{} & 1.00 & 26 & 21 & 0 slots & 121463.57 \\
GPT & \textbf{\compfrozen{}} & \textbf{1.00} & \textbf{1} & \textbf{0} & \textbf{5 slots} & \textbf{4356.69} \\
\midrule
Gemini & Comp-EntryOnly & 0.00 & 0 & 0 & 5 slots & 1.33 \\
Gemini & \retryonly{} & 1.00 & 26 & 21 & 0 slots & 74006.58 \\
Gemini & \textbf{\compfrozen{}} & \textbf{1.00} & \textbf{1} & \textbf{0} & \textbf{5 slots} & \textbf{2182.24} \\
\midrule
DeepSeek & Comp-EntryOnly & 0.00 & 0 & 0 & 5 slots & 1.46 \\
DeepSeek & \retryonly{} & 1.00 & 26 & 21 & 0 slots & 141557.97 \\
DeepSeek & \textbf{\compfrozen{}} & \textbf{1.00} & \textbf{1} & \textbf{0} & \textbf{5 slots} & \textbf{5372.30} \\
\midrule
Qwen & Comp-EntryOnly & 0.00 & 0 & 0 & 5 slots & 0.82 \\
Qwen & \retryonly{} & 1.00 & 26 & 21 & 0 slots & 45189.98 \\
Qwen & \textbf{\compfrozen{}} & \textbf{1.00} & \textbf{1} & \textbf{0} & \textbf{5 slots} & \textbf{1614.10} \\
\bottomrule
\end{tabular}
\end{adjustbox}
\end{table*}

\section{Cross-Model Generalization Results}
\label{app:xmodel-generalization}

Appendix Table~\ref{tab:xmodel-generalization} reports the full cross-model generalization results beyond the decisive row, covering cross-domain, cross-runtime, and control-case settings.

\begin{table*}[!htbp]
\caption{Cross-model generalization results beyond the decisive row.
We report representative results from three broader settings: a navigation commitment-sensitive row (cross-domain evidence), a LangGraph-based schedule row (cross-runtime evidence), and an official schedule-form control case (control-case evidence). Within each setting, we report a single representative recovery controller per model to keep cross-model comparisons aligned. Columns are ordered by priority from left to right: success, replay width, upstream replay, preserved progress, and failure-to-milestone latency.}
\label{tab:xmodel-generalization}
\centering
\scriptsize
\setlength{\tabcolsep}{3pt}
\begin{adjustbox}{max width=\textwidth}
\begin{tabular}{lllccccc}
\toprule
Setting & Model & Method & Success & Replay & Up. replay & Preserved & F$\rightarrow$M (ms) \\
\midrule
Navigation row & GLM & \compfrozen{} & 1.00 & 1 & 0 & 1 stop & 1733.45 \\
Navigation row & GPT & \compfrozen{} & 0.80 & 1 & 0 & 1 stop & 7445.63 \\
Navigation row & Gemini & \compfrozen{} & 1.00 & 1 & 0 & 1 stop & 3894.71 \\
Navigation row & DeepSeek & \compfrozen{} & 1.00 & 1 & 0 & 1 stop & 12718.58 \\
Navigation row & Qwen & \compfrozen{} & 1.00 & 1 & 0 & 1 stop & 5282.35 \\
\midrule
LangGraph row & GLM & LangGraph-SemiReal & 1.00 & 1 & 0 & 5 slots & 1038.74 \\
LangGraph row & GPT & LangGraph-SemiReal & 1.00 & 1 & 0 & 5 slots & 4310.66 \\
LangGraph row & Gemini & LangGraph-SemiReal & 1.00 & 1 & 0 & 5 slots & 2489.38 \\
LangGraph row & DeepSeek & LangGraph-SemiReal & 1.00 & 1 & 0 & 5 slots & 4184.86 \\
LangGraph row & Qwen & LangGraph-SemiReal & 1.00 & 1 & 0 & 5 slots & 1805.62 \\
\midrule
Official control & GLM & \compfrozen{} & 1.00 & 8 & 0 & 3 slots & 8093.18 \\
Official control & GPT & \compfrozen{} & 1.00 & 8 & 0 & 3 slots & 10756.99 \\
Official control & Gemini & \compfrozen{} & 1.00 & 8 & 0 & 3 slots & 11477.65 \\
Official control & DeepSeek & \compfrozen{} & 1.00 & 8 & 0 & 3 slots & 10647.14 \\
Official control & Qwen & \compfrozen{} & 1.00 & 8 & 0 & 3 slots & 10046.28 \\
\bottomrule
\end{tabular}
\end{adjustbox}
\end{table*}

\section{Benchmark Universe and Deterministic-Domain Generalization}
\label{app:generalization}

Appendix~\ref{app:generalization} broadens the empirical scope beyond the three main-text LLM-driven domains by documenting the full five-domain benchmark universe and the deterministic-domain generalization results. The travel-planning domain and its included cases are derived from the open-source TravelPlanner dataset~\citep{xie2024travelplanner}.

\subsection{Five-Domain Benchmark Universe}

\begin{table}[!htbp]
\caption{Five-domain benchmark universe. The main text uses the three core LLM-driven domains; deterministic ETL and travel-planning results appear only in Appendix~\ref{app:generalization}. All live aggregates use repeat~=~5.}
\label{tab:appendix-case-universe}
\centering
\scriptsize
\setlength{\tabcolsep}{3pt}
\begin{adjustbox}{max width=\linewidth}
\begin{tabular}{llrrrcc}
\toprule
Domain & Role & Official cases & Commit-sensitive cases & Repeat & LLM-driven & Deterministic \\
\midrule
Navigation & main text & 4 & 8 & 5 & yes & no \\
Schedule Form & main text & 4 & 7 & 5 & yes & no \\
Diagnosis & main text & 4 & 6 & 5 & yes & no \\
ETL Pipeline & appendix & 7 & 7 & 5 & no & yes \\
Travel Planning & appendix & 4 & 3 & 5 & no & yes \\
\bottomrule
\end{tabular}
\end{adjustbox}
\end{table}

The five-domain universe is intentionally heterogeneous. Navigation, schedule-form, and diagnosis are externally grounded LLM-driven live-agent domains. ETL pipeline and travel planning are deterministic domains that stress the same recovery framework without live LLM uncertainty, so we keep them as appendix-only generalization evidence. Travel-planning cases instantiate TravelPlanner tasks as frozen planning skeletons with explicit commit/exit predicates and controlled observable-failure sites, while audit-safe-equivalence is still evaluated with frozen domain specifications.

\subsection{Deterministic-Domain Generalization}

\begin{table*}[!htbp]
\caption{Deterministic-domain generalization on ETL pipeline and travel planning. These rows use the same stronger-baseline protocol as the main text; travel-planning cases come from TravelPlanner~\citep{xie2024travelplanner}. Status uses the same semantics as Table~\ref{tab:recovery-main}. For compactness, preserved completed instances is omitted here and the table focuses on task-level success and replay behavior in deterministic domains. Values are medians over successful runs.}
\label{tab:appendix-deterministic-generalization}
\centering
\scriptsize
\setlength{\tabcolsep}{3pt}
\begin{adjustbox}{max width=\textwidth}
\begin{tabular}{lllrrrrrr}
\toprule
Regime & Domain & Method & Success & Replay & Upstream replay & Failure$\rightarrow$milestone (ms) & Recov. obs. & Status \\
\midrule
Official & ETL Pipeline & \retryonly{} & 1.00 & 10.0 & 5.0 & 0.89 & 0.00 & ok \\
Official & ETL Pipeline & Comp-EntryOnly & 1.00 & 4.0 & 0.0 & 4.53 & 1.00 & ok \\
Official & ETL Pipeline & \textbf{\compfrozen{}} & \textbf{1.00} & \textbf{4.0} & \textbf{0.0} & \textbf{4.50} & \textbf{1.00} & \textbf{ok} \\
Official & Travel Planning & \retryonly{} & 1.00 & 7.0 & 0.0 & 0.50 & 0.00 & ok \\
Official & Travel Planning & Comp-EntryOnly & 1.00 & 1.0 & 0.0 & 0.47 & 1.00 & ok \\
Official & Travel Planning & \textbf{\compfrozen{}} & \textbf{1.00} & \textbf{1.0} & \textbf{0.0} & \textbf{0.47} & \textbf{1.00} & \textbf{ok} \\
\midrule
Commit-sensitive & ETL Pipeline & \retryonly{} & 1.00 & 22.0 & 15.0 & 1.71 & 0.00 & ok \\
Commit-sensitive & ETL Pipeline & Comp-EntryOnly & 0.00 & -- & -- & -- & 0.00 & no-recov \\
Commit-sensitive & ETL Pipeline & \textbf{\compfrozen{}} & \textbf{1.00} & \textbf{1.0} & \textbf{0.0} & \textbf{18.58} & \textbf{1.00} & \textbf{ok} \\
Commit-sensitive & Travel Planning & \retryonly{} & 1.00 & 17.0 & 0.0 & 0.61 & 0.00 & ok \\
Commit-sensitive & Travel Planning & Comp-EntryOnly & 0.00 & -- & -- & -- & 0.00 & blocked \\
Commit-sensitive & Travel Planning & \textbf{\compfrozen{}} & \textbf{1.00} & \textbf{1.0} & \textbf{0.0} & \textbf{0.63} & \textbf{1.00} & \textbf{ok} \\
\bottomrule
\end{tabular}
\end{adjustbox}
\end{table*}

The deterministic domains support a narrower generalization claim: in the official setting, \compfrozen{} continues to eliminate upstream replay, and in the deterministic commitment-sensitive settings entry-only recovery again fails whereas \compfrozen{} still succeeds with a one-step replay frontier.

\section{Statistical Robustness and Efficiency}
\label{app:stats}

Appendix~\ref{app:stats} complements the main-text tables with paired significance tests, no-failure-path efficiency measurements, and a compact checkpoint-granularity diagnostic.

\subsection{Paired Live Statistics}
\label{app:paired-live-robustness}

\begin{table*}[!htbp]
\caption{Paired latency robustness for \compfrozen{} versus \retryonly{}. Rows use runtime-exported pair keys and report matched-run medians, paired median deltas with 95\% bootstrap confidence intervals, and Holm-adjusted exact tests. Replay and upstream replay also uniformly favor \compfrozen{}.}
\label{tab:paired-live-robustness}
\centering
\scriptsize
\setlength{\tabcolsep}{3pt}
\begin{adjustbox}{max width=\textwidth}
\begin{tabular}{llrrrrr}
\toprule
Regime & Domain & Paired $\Delta$ [95\% CI] & Effect / Holm-$p$ & \retryonly{} median (ms) & \compfrozen{} median (ms) & Pairs \\
\midrule
Official & Navigation & -3202.20 [-3744.90, -1939.54] & $d_z=-1.46$, $7.63\times 10^{-5}$ & 6920.20 & 4841.55 & 20 \\
Official & Schedule Form & -12962.77 [-14470.23, -11189.84] & $d_z=-1.88$, $7.63\times 10^{-6}$ & 18530.74 & 4431.35 & 20 \\
Official & Diagnosis & -6916.97 [-7173.38, -6343.63] & $d_z=-2.33$, $7.63\times 10^{-6}$ & 12577.31 & 5736.50 & 20 \\
\midrule
Commit-sensitive & Navigation & -21825.54 [-25423.12, -17380.94] & $d_z=-1.68$, $5.00\times 10^{-5}$ & 25752.95 & 2527.10 & 40 \\
Commit-sensitive & Schedule Form & -33867.34 [-34671.56, -32231.64] & $d_z=-4.85$, $5.00\times 10^{-5}$ & 34690.47 & 1141.17 & 35 \\
Commit-sensitive & Diagnosis & -19142.11 [-20507.09, -16356.22] & $d_z=-2.77$, $5.00\times 10^{-5}$ & 21174.59 & 2113.80 & 30 \\
\bottomrule
\end{tabular}
\end{adjustbox}
\end{table*}

The paired statistics confirm the main text's qualitative pattern. The biggest latency reductions appear in the commitment-sensitive rows, and the official headline setting remains significant in all three core LLM-driven domains when compared against \retryonly{}. Against stronger local baselines on official rows, however, \compfrozen{} does not show uniform dominance: in navigation it is statistically indistinguishable from both Coarse-State-Retry and Comp-EntryOnly on failure-to-milestone latency (Holm-adjusted $p=0.18352$ and $1.00000$), with replay, upstream replay, and preserved completed instances all at parity; in schedule-form it retains selective latency improvements over both Coarse-State-Retry and Comp-EntryOnly (Holm-adjusted $p=0.00060$ and $0.00616$), while replay, upstream replay, and preserved completed instances remain at parity.

\subsection{No-Failure-Path Overhead}
\label{app:no-failure-overhead}

\begin{table*}[!htbp]
\caption{No-failure-path overhead for the currently instrumented navigation and schedule-form domains. Values are medians over repeat~=~5. The sidecar-hook and snapshot columns isolate recovery-readiness bookkeeping rather than total planner variance.}
\label{tab:no-failure-overhead}
\centering
\scriptsize
\setlength{\tabcolsep}{3pt}
\begin{adjustbox}{max width=\textwidth}
\begin{tabular}{llrrrrr}
\toprule
Domain & Method & Sidecar hook (ms) & Snapshot subset (ms) & Peak snapshot bytes & Total wall-clock (ms) & Planner (ms) \\
\midrule
Navigation & \retryonly{} & 0.00 & 0.00 & 0 & 8131.18 & 4558.15 \\
Navigation & \textbf{\compfrozen{}} & 19.47 & 13.31 & 15072 & 9298.71 & 4413.66 \\
Schedule Form & \retryonly{} & 0.00 & 0.00 & 0 & 27014.20 & 26997.02 \\
Schedule Form & \textbf{\compfrozen{}} & 46.19 & 34.37 & 12353 & 29918.02 & 29853.67 \\
\bottomrule
\end{tabular}
\end{adjustbox}
\end{table*}

Even though total no-failure wall-clock still reflects live-path variance, the bookkeeping attributable to \compfrozen{} remains small in absolute terms: the sidecar hook stays on the order of tens of milliseconds, and peak serialized snapshots remain around 12--15\,KB in the current setup.

\subsection{Checkpoint Granularity Diagnostic}
\label{app:granularity}

\begin{table}[!htbp]
\caption{Synthetic checkpoint-granularity diagnostic in navigation. Once the admissible restore point lies beyond entry, reviewed commit checkpoints sharply reduce replay. End-to-end latency is a synthetic harness estimate.}
\label{tab:granularity}
\centering
\scriptsize
\setlength{\tabcolsep}{3.5pt}
\begin{adjustbox}{max width=\linewidth}
\begin{tabular}{lrrr}
\toprule
Method & Avg replay steps & Avg upstream replay & Avg estimated end-to-end latency (ms) \\
\midrule
\retryonly{} & 12.0 & 8.0 & 8209.75 \\
Comp-EntryOnly & 4.0 & 0.0 & 3751.00 \\
\textbf{\compfrozen{}} & \textbf{1.0} & \textbf{0.0} & \textbf{872.00} \\
\bottomrule
\end{tabular}
\end{adjustbox}
\end{table}

\section{External Validation on LangGraph-Based Runtimes}
\label{app:langgraph}

Appendix~\ref{app:langgraph} provides \emph{external validation} of the main recoverability claim on LangGraph-based runtimes. We instantiate two LangGraph-based recovery controllers and evaluate them through three complementary views: regime-aware comparison, transplant-control transportability, and a counterfactual blocking witness. The goal is not to benchmark generic framework speed, but to test whether the same \emph{commitment-sensitive rollback failure} isolated by the admissibility analysis reappears once persistence and resume behavior are already available in an external graph runtime; this two-domain overlay is external evidence only and is not part of the five-domain audit denominator.

\subsection{Runtime Families}

\begin{table}[!htbp]
\caption{Runtime families in the external validation study. LangGraph-Direct and LangGraph-SemiReal are implemented using LangGraph execution and persistence primitives under our controlled benchmark protocol.}
\label{tab:langgraph-runtime-families}
\centering
\scriptsize
\setlength{\tabcolsep}{3pt}
\begin{adjustbox}{max width=\linewidth}
\begin{tabular}{llll}
\toprule
Runtime family & Primary use & Realization & Scope \\
\midrule
Direct LangGraph-based & sanity baseline & LangGraph-Direct & aligned 2-case/domain subset, repeat=5 \\
Semi-real LangGraph-based & strong external baseline & LangGraph-SemiReal & official and regime-balanced 4-case/domain sets, repeat=5 \\
Proposed runtime & recoverability controller & DART & matched recovery protocol across aligned comparisons \\
\bottomrule
\end{tabular}
\end{adjustbox}
\end{table}

\subsection{Regime-Aware Comparison}

Readers interested mainly in the decisive external result may start from Table~\ref{tab:langgraph-three-way-anchor}, where the schedule-form commitment-sensitive row isolates the boundary failure predicted by the main admissibility analysis. The aggregate rows here provide context: on the smaller aligned direct subset, LangGraph-Direct is faster on raw failure-to-milestone time in both headline domains (navigation: 1480.22 ms vs.\ 3248.19 ms; schedule-form: 1.77 ms vs.\ 2426.61 ms). The decisive difference therefore lies in the schedule-form commitment-sensitive regime, where rollback admissibility becomes outcome-critical.

\begin{table*}[!htbp]
\caption{Aggregate LangGraph-SemiReal vs.\ DART summaries on the official and regime-balanced aligned sets. Values are medians over successful runs.}
\label{tab:langgraph-summary}
\centering
\scriptsize
\setlength{\tabcolsep}{3pt}
\begin{adjustbox}{max width=\textwidth}
\begin{tabular}{llrrrrrr}
\toprule
Track & Domain & LangGraph success & DART success & LangGraph replay & DART replay & LangGraph F$\rightarrow$M (ms) & DART F$\rightarrow$M (ms) \\
\midrule
Official & Navigation & 1.00 & 1.00 & 1.0 & 5.5 & 4.33 & 9619.25 \\
Official & Schedule Form & 1.00 & 1.00 & 1.5 & 4.5 & 3148.41 & 5589.87 \\
Regime-balanced & Navigation & 1.00 & 1.00 & 1.0 & 2.0 & 1283.99 & 3137.55 \\
Regime-balanced & Schedule Form & 1.00 & 1.00 & 1.5 & 3.0 & 3629.59 & 7421.76 \\
\bottomrule
\end{tabular}
\end{adjustbox}
\end{table*}

The regime-aware three-way anchor is the clearest external-validation result because it keeps only the shared regime-specific case intersection across \retryonly{}, LangGraph-SemiReal, and \dart{} \compfrozen{}. On navigation, the semi-real LangGraph runtime remains competitive in both entry-aligned and commitment-sensitive settings. On schedule-form, however, the commitment-sensitive row is decisive: \retryonly{} succeeds with a large replay frontier (32618.41 ms, 25.5 replayed actions), LangGraph-SemiReal drops to 0.00 success, and \dart{} remains admissible and succeeds with a one-step replay frontier (1109.80 ms, 1.0 replayed action). This is the external confirmation that checkpoint-aligned restore alone is insufficient once rollback must respect downstream commitments.

To separate policy effects from executor effects, we also run a matched G0/G1/G2/G3 transplant-control study on the same two-domain regime-balanced universe while holding the reviewed checkpoint substrate fixed between G2 and G3. In all four domain-regime cells, G2 and G3 both remain at 1.00 success with identical median replay and zero upstream replay; on the current universe every G3 decision is \texttt{eligible} with zero fallbacks, and relative to DART-native the transplanted controller matches replay exactly while keeping failure-to-milestone latency within $0.76$--$0.84\times$. These rows are therefore a transportability check rather than blocked-case evidence: they show that the admissibility layer ports without harming recovery on safe cases. Because every current G3 decision is \texttt{eligible}, the necessity of the gate comes instead from the counterfactual witness below and the five-domain calibration results in Appendix~\ref{app:audit}, which show what breaks when the same dependency/effect veto is disabled or overridden.

\begin{table*}[!htbp]
\caption{Three-way anchor across \retryonly{}, LangGraph-SemiReal, and DART \compfrozen{} on aligned regime-specific intersections. The schedule-form commitment-sensitive row is the key safety result.}
\label{tab:langgraph-three-way-anchor}
\centering
\scriptsize
\setlength{\tabcolsep}{3pt}
\begin{adjustbox}{max width=\textwidth}
\begin{tabular}{llrrrrrrrrr}
\toprule
Domain & Regime & Retry success & LangGraph success & DART success & Retry replay & LangGraph replay & DART replay & Retry F$\rightarrow$M & LangGraph F$\rightarrow$M & DART F$\rightarrow$M \\
\midrule
Navigation & Entry-aligned & 1.00 & 1.00 & 1.00 & 10.0 & 1.5 & 4.0 & 6376.97 & 588.89 & 3328.69 \\
Navigation & Commit-sensitive & 1.00 & 1.00 & 1.00 & 12.0 & 1.0 & 1.0 & 14596.55 & 1938.46 & 2433.90 \\
Schedule Form & Entry-aligned & 1.00 & 1.00 & 1.00 & 21.5 & 4.0 & 6.5 & 19093.78 & 4969.47 & 6137.15 \\
Schedule Form & Commit-sensitive & 1.00 & 0.00 & 1.00 & 25.5 & 0.0 & 1.0 & 32618.41 & -- & 1109.80 \\
\bottomrule
\end{tabular}
\end{adjustbox}
\end{table*}

\subsection{Semantic Equivalence}

The study is not relying only on latency. Under the current overlay contract, all comparable LangGraph-SemiReal versus \dart{} pairs are safe-equivalent: 6/6 on the official track and 8/8 on the regime-balanced track.

\begin{table}[!htbp]
\caption{LangGraph-SemiReal semantic overlays. The official overlay yields 6 comparable safe-equivalent rows, and the regime-balanced overlay yields 8.}
\label{tab:langgraph-semantic-overlay}
\centering
\scriptsize
\setlength{\tabcolsep}{3pt}
\begin{adjustbox}{max width=\linewidth}
\begin{tabular}{lrrrr}
\toprule
Track & Safe-equivalent rows & Comparable rows & Navigation rows & Schedule-form rows \\
\midrule
Official & 6 & 6 & 2 & 4 \\
Regime-balanced & 8 & 8 & 4 & 4 \\
\bottomrule
\end{tabular}
\end{adjustbox}
\end{table}

\subsection{Mechanism Witness}

The blocking witness is counterfactual evidence that the gate is necessary, not just that \dart{} differs from LangGraph. In the schedule-form witness, producer \texttt{ResolveSlot::slot[0]::0} has already been consumed by two committed downstream instances: \texttt{ResolveSlot::slot[1]::0} and \texttt{FinalizeSchedule::final::0}. With blocking on, \dart{} rejects rollback with \texttt{committed\_consumers\_present}. With the same check disabled, the producer-commit restore becomes \texttt{eligible} but drops those two committed consumers. The reconstructed LangGraph-style restore invalidates the same pair. So checkpoint alignment alone is not enough once rollback must preserve downstream commitments.

\begin{table}[!htbp]
\caption{Schedule-form blocking witness against LangGraph-SemiReal under an explicit producer rollback request. Without blocking, the same restore invalidates committed downstream consumers.}
\label{tab:langgraph-blocking-witness}
\centering
\scriptsize
\setlength{\tabcolsep}{2pt}
\begin{adjustbox}{max width=\linewidth}
\begin{tabular}{>{\raggedright\arraybackslash}p{0.20\linewidth}>{\raggedright\arraybackslash}p{0.20\linewidth}>{\raggedright\arraybackslash}p{0.24\linewidth}>{\raggedright\arraybackslash}p{0.24\linewidth}}
\toprule
Setting & DART blocking on & DART blocking off & LangGraph-SemiReal restore \\
\midrule
Schedule producer rollback after finalized downstream consumers & reject with \texttt{committed\_consumers\_} \newline \texttt{present} & rollback allowed; \newline 2 committed consumers dropped & restore allowed; \newline same 2 downstream consumers invalidated \\
\bottomrule
\end{tabular}
\end{adjustbox}
\end{table}

\section{Five-Domain Audit Chain and Mechanism Evidence}
\label{app:audit}

Appendix~\ref{app:audit} collects the five-domain audit chain and the mechanism-level sanity checks that underpin the main-text semantic audit and blocking calibration claims, including semantic audit, blocking calibration, failed-instance localization, and property-wise necessity evidence. Reviewers interested only in the main-text safety claims can read this appendix in the following order: denominator flow and protocol, five-domain semantic audit, five-domain blocking calibration, and then the localization and boundary/property evidence.

\subsection{Five-Domain Semantic Audit and Blocking Calibration}

These tables establish the safety backbone behind Table~\ref{tab:correctness-safety}. Table~\ref{tab:denominator-flow} aligns the main counts across the five-domain audit chain, separating the comparable-row denominator used for safe-equivalence from the evaluated-event denominator used for blocking calibration. The reported labels are executable checks derived from frozen case specifications and reviewed domain specifications over normalized semantic, committed-prefix, and durable-effect projections. Human review enters when freezing boundary, effect, and audit specifications; we do not claim an independent multi-annotator audit for the current version.

\begin{table}[!htbp]
\caption{Denominator flow for the five-domain audit chain.}
\label{tab:denominator-flow}
\centering
\scriptsize
\setlength{\tabcolsep}{3pt}
\begin{adjustbox}{max width=\linewidth}
\begin{tabular}{p{0.34\linewidth}r p{0.44\linewidth}}
\toprule
Denominator & Count & Source / scope \\
\midrule
Comparable semantic rows & 54 & Five-domain frozen audit universe \\
Evaluated recovery events & 47 & Calibration profile over the same universe \\
Admitted events & 35 & Subset of evaluated recovery events \\
Blocked events & 12 & Subset of evaluated recovery events \\
Blocked checkpoints & 16 & Audit-level blocked-checkpoint total \\
Repeat-level localization rows & 270 & 54 frozen cases $\times$ 5 repeats \\
Ambiguity candidates & 291 & Offline weakened-key benchmark \\
\bottomrule
\end{tabular}
\end{adjustbox}
\end{table}

The 54-row semantic denominator counts cases where both methods yield audit-ready terminal outputs; for injected-failure runs, comparability additionally requires observed failure and recovery under \compfrozen{}. The 47-event calibration denominator is the subset that reaches the admissibility gate. Shared status labels are \texttt{ok}, \texttt{contract}, \texttt{no-recov}, and \texttt{blocked}, with the same meanings as in Table~\ref{tab:recovery-main}. We use \emph{admitted event}, \emph{blocked event}, and \emph{blocked checkpoint} for the three calibration units, and reserve \emph{preserved completed instances} for the runtime metric versus \emph{committed-prefix preservation} for the ETL/travel audit check. A comparable row is safe-equivalent iff all domain-applicable reviewed checks pass; rows failing the comparable-row inclusion rule are excluded from the safe-equivalence denominator.

\begin{table*}[!htbp]
\caption{Audit protocol and denominators for Table~\ref{tab:correctness-safety}.}
\label{tab:appendix-audit-protocol}
\centering
\scriptsize
\setlength{\tabcolsep}{3pt}
\begin{adjustbox}{max width=\textwidth}
\begin{tabular}{>{\raggedright\arraybackslash}p{0.19\textwidth}>{\raggedright\arraybackslash}p{0.18\textwidth}>{\raggedright\arraybackslash}p{0.17\textwidth}>{\raggedright\arraybackslash}p{0.24\textwidth}>{\raggedright\arraybackslash}p{0.10\textwidth}}
\toprule
Quantity & Current result & Reported denominator & Pass / failure rule & Unit \\
\midrule
Semantic audit & 54/54 safe-equivalent & 54 comparable rows & safe-equivalent iff all required frozen specifications pass & comparable row \\
False blocking (event level) & 0/12 false-blocked events & 12 audited blocked events & false-blocked iff forced restore remains audit-safe & blocked event \\
False blocking (checkpoint level) & 0/16 false-blocked checkpoints & 16 audited blocked checkpoints & false-blocked iff the blocked checkpoint itself is safe to force & blocked checkpoint \\
Unsafe admission & 0/35 unsafe admissions & 35 admitted events & unsafe iff any required reviewed specification fails after admission & admitted event \\
\bottomrule
\end{tabular}
\end{adjustbox}
\end{table*}

\begin{table*}[!htbp]
\caption{Domain-specific reviewed specifications for audit-safe-equivalence.}
\label{tab:appendix-audit-specifications}
\centering
\scriptsize
\setlength{\tabcolsep}{3pt}
\begin{adjustbox}{max width=\textwidth}
\begin{tabular}{p{0.14\textwidth}p{0.45\textwidth}p{0.31\textwidth}}
\toprule
Domain & Safe-equivalent requirement & Comparable-row requirement \\
\midrule
Navigation & canonical final semantics + unaffected committed prefix & both methods yield auditable terminal outputs; \compfrozen{} must observe the injected failure and recovery when expected \\
Schedule Form & canonical final semantics + unaffected committed prefix & both methods yield auditable terminal outputs; \compfrozen{} must observe the injected failure and recovery when expected \\
Diagnosis & canonical final semantics + unaffected committed prefix + durable repair effect & both methods complete the diagnosis probe outputs on the same case \\
ETL Pipeline & canonical final semantics + unaffected committed prefix + durable effect + committed-prefix preservation & both methods complete the ETL probe outputs on the same case \\
Travel Planning & canonical final semantics + unaffected committed prefix + durable effect + committed-prefix preservation & both methods complete the travel-planning probe outputs on the same case \\
\bottomrule
\end{tabular}
\end{adjustbox}
\end{table*}

\begin{table}[!htbp]
\caption{Five-domain semantic audit underlying Table~\ref{tab:correctness-safety}. Rows are aggregated at the comparable-row level.}
\label{tab:appendix-semantic-five-domain}
\centering
\scriptsize
\setlength{\tabcolsep}{3pt}
\begin{adjustbox}{max width=\linewidth}
\begin{tabular}{lrrrrrr}
\toprule
Domain & Safe-equivalent & Comparable rows & Semantic match & Prefix exact & Effect exact & Committed-prefix exact \\
\midrule
Navigation & 12 & 12 & 12 & 12 & 0 & 0 \\
Schedule Form & 11 & 11 & 11 & 11 & 0 & 0 \\
Diagnosis & 10 & 10 & 10 & 10 & 10 & 0 \\
ETL Pipeline & 14 & 14 & 14 & 14 & 14 & 14 \\
Travel Planning & 7 & 7 & 7 & 7 & 7 & 7 \\
\midrule
Overall & 54 & 54 & 54 & 54 & 31 & 21 \\
\bottomrule
\end{tabular}
\end{adjustbox}
\end{table}

\begin{table}[!htbp]
\caption{Five-domain blocking calibration by domain. Rows use the same 47 evaluated recovery events as Table~\ref{tab:correctness-safety}; false-blocked and unsafe-admission columns report audited event-level counts.}
\label{tab:appendix-blocking-five-domain}
\centering
\scriptsize
\setlength{\tabcolsep}{3pt}
\begin{adjustbox}{max width=\linewidth}
\begin{tabular}{lrrrrrrr}
\toprule
Scope & Admitted & Blocked & False-blocked & Unsafe admissions & Dependency-blocked & Effect-blocked & Evaluated events \\
\midrule
Overall calibration & 35 & 12 & 0 & 0 & 5 & 7 & 47 \\
Navigation & 4 & 1 & 0 & 0 & 1 & 0 & 5 \\
Schedule Form & 4 & 2 & 0 & 0 & 1 & 1 & 6 \\
Diagnosis & 6 & 2 & 0 & 0 & 1 & 1 & 8 \\
ETL Pipeline & 14 & 3 & 0 & 0 & 1 & 2 & 17 \\
Travel Planning & 7 & 4 & 0 & 0 & 1 & 3 & 11 \\
\bottomrule
\end{tabular}
\end{adjustbox}
\end{table}

\subsection{Failed-Instance Localization Audit}
\label{app:localization-audit}

This subsection audits failed-instance localization over the same five-domain frozen case universe defined in Table~\ref{tab:appendix-case-universe} (54 cases, 5 repeats each, 270 repeat-level rows). It addresses a narrower but reviewer-critical question than semantic equivalence alone: whether the runtime actually localizes recovery to the correct failed instance rather than merely to the correct skeleton family. We audit this in three layers. First, we align the observed \texttt{Comp-Frozen} recovery scope and checkpoint type with frozen case specifications over the repeat~=~5 official and commitment-sensitive cases. Across the 270 repeat-level rows, the observed recovery-scope prefix matches the specification in all rows, and the observed checkpoint type also matches in all rows. Second, we run a systematic offline ambiguity benchmark over the same universe: the benchmark keeps the observed full recovery identifiers fixed, then weakens instance keys by dropping ordinal or structural fields to test when a conservative runtime should abstain. Third, we retain three executable consequence probes. Navigation and diagnosis each yield a unique weakened-alias candidate under the frozen protocol, whereas schedule-form exposes a genuine re-entry ambiguity witness: collapsing ordinal identity creates two candidates, and forcing the stale one would erase an already committed refined value. The probes therefore illustrate concrete ambiguity damage, while the broader ambiguity benchmark provides the systematic coverage.

\begin{table}[!htbp]
\caption{Failed-instance localization audit. The table combines observed repeat-level alignment, a systematic offline ambiguity benchmark over the frozen case universe, and three executable consequence probes.}
\label{tab:appendix-localization-audit}
\centering
\scriptsize
\setlength{\tabcolsep}{2pt}
\begin{adjustbox}{max width=\linewidth}
\begin{tabular}{>{\raggedright\arraybackslash}p{0.28\linewidth}>{\raggedright\arraybackslash}p{0.24\linewidth}>{\raggedright\arraybackslash}p{0.24\linewidth}>{\raggedright\arraybackslash}p{0.14\linewidth}}
\toprule
Component & Key result & Interpretation & Denominator \\
\midrule
Observed repeat-level alignment & recovery-scope prefix: 270/270; checkpoint type: 270/270 & observed \texttt{Comp-Frozen} rows align with the frozen recovery specifications & 270 rows \\
Systematic ambiguity benchmark & full-key exact: 270/270 observed rows; drop-ordinal ambiguity: 20/291 candidate instances & full instance identifiers remain unique, whereas weakened identifiers require abstention in re-entry patterns & 54 cases / 291 candidates \\
Re-entry stress aliases & same-entity multi-ordinal aliases appear in navigation and schedule-form & ordinal identity becomes necessary once the same structural entity is reopened & 10 aliases \\
Consequence probes & navigation unique; schedule ambiguous with abstention; diagnosis unique & the probes illustrate concrete ambiguity damage rather than defining the full benchmark & 3 probes \\
Overall localization takeaway & no false resolution observed under full keys in the current frozen universe & locality claims are supported over the reviewed five-domain case universe rather than arbitrary online executions & combined audit \\
\bottomrule
\end{tabular}
\end{adjustbox}
\end{table}

\subsection{Boundary Review Protocol and Boundary Evidence}
\label{app:boundary-audit}

\begin{table*}[!htbp]
\caption{Reviewed positive exit-boundary cases and paired negative controls in the two core workflow-shaped domains. Together with Table~\ref{tab:mechanism-ablation}, these rows show that controller legality is weaker than reviewed boundary validity.}
\label{tab:boundary-evidence}
\centering
\scriptsize
\setlength{\tabcolsep}{3pt}
\begin{adjustbox}{max width=\textwidth}
\begin{tabular}{llllll}
\toprule
Domain & Edge & Reviewed exit boundary & Planning/runtime realized & Conclusion & FSM legal \\
\midrule
Navigation & \texttt{WAITING\_QUERY\_REFINEMENT$\rightarrow$STOP\_READY} & yes & yes & exit boundary & yes \\
Navigation & \texttt{WAITING\_QUERY\_REFINEMENT$\rightarrow$ROUTES\_PLANNED} & yes & yes & exit boundary & yes \\
Navigation & \texttt{WAITING\_QUERY\_REFINEMENT$\rightarrow$DONE} & yes & yes & exit boundary & yes \\
Navigation & \texttt{WAITING\_POI\_SELECTION$\rightarrow$STOP\_READY} & no & no & legal but non-recoverable & yes \\
Schedule Form & \texttt{WAITING\_SLOT\_REFINEMENT$\rightarrow$SLOT\_READY} & yes & yes & exit boundary & yes \\
Schedule Form & \texttt{WAITING\_SLOT\_REFINEMENT$\rightarrow$REVIEW\_READY} & yes & yes & exit boundary & yes \\
Schedule Form & \texttt{WAITING\_SLOT\_REFINEMENT$\rightarrow$DONE} & yes & yes & exit boundary & yes \\
Schedule Form & \texttt{WAITING\_SLOT\_SELECTION$\rightarrow$SLOT\_READY} & no & no & legal but non-recoverable & yes \\
\bottomrule
\end{tabular}
\end{adjustbox}
\end{table*}

\begin{table}[!htbp]
\caption{Boundary review load across workflow-shaped reviewed-boundary domains. Counts report reviewed boundary objects rather than full FSM size; ETL pipeline and travel planning instead use deterministic output/effect specifications and are outside this reviewer-burden view.}
\label{tab:boundary-audit}
\centering
\scriptsize
\setlength{\tabcolsep}{3pt}
\begin{adjustbox}{max width=\linewidth}
\begin{tabular}{lrrrrrrl}
\toprule
Domain & \#Skeletons & \#Commit & \#Exit & \#Pending & \#States & \#Legal edges & Audit target \\
\midrule
Navigation & 3 & 4 & 6 & 1 & 14 & 28 & candidate predicates + interface/effect audit \\
Schedule Form & 3 & 4 & 6 & -- & 12 & 18 & candidate predicates + interface/effect audit \\
Diagnosis & 3 & 4 & 6 & -- & 13 & 19 & candidate predicates + interface/effect audit \\
\bottomrule
\end{tabular}
\end{adjustbox}
\end{table}

\begin{table}[!htbp]
\caption{Protocolized boundary-review telemetry from logged real sessions in a representative workflow-shaped domain. Reviewer-minute quantities are reported only when a real timed session is present; exact match checks whether the reviewed output reproduces the frozen configuration after validation.}
\label{tab:boundary-review-telemetry}
\centering
\scriptsize
\setlength{\tabcolsep}{3pt}
\begin{adjustbox}{max width=\linewidth}
\begin{tabular}{l l r r r r r r}
\toprule
Domain & Reviewer role & Frozen config (min) & First config (min) & Manual decisions & Validation iters & Exact match & Sessions \\
\midrule
Schedule Form & author (not original implementer) & 27.0 & 18.5 & 4 & 2 & 1.00 & 1 \\
\bottomrule
\end{tabular}
\end{adjustbox}
\end{table}

\begin{table}[!htbp]
\caption{Cross-domain structural boundary transfer audit from candidate export to frozen configuration. Stable indicates no missing or extra skeletons and no field diffs relative to the frozen reviewed configuration; Exact match refers to candidate-to-frozen alignment at the skeleton/field level.}
\label{tab:boundary-transfer}
\centering
\scriptsize
\setlength{\tabcolsep}{3pt}
\begin{adjustbox}{max width=\linewidth}
\begin{tabular}{l l r r r r}
\toprule
Domain & Exact match & Stable & Field diffs & Pending & Candidate source \\
\midrule
Navigation & yes & yes & 0 & 1 & FSM+router+manifest \\
Schedule Form & yes & yes & 0 & 0 & FSM+planner-map+manifest \\
Diagnosis & yes & yes & 0 & 0 & FSM+planner-map+manifest \\
\bottomrule
\end{tabular}
\end{adjustbox}
\end{table}

These tables make the scope explicit: reviewer-burden evidence is reported only for workflow-shaped reviewed-boundary domains, with timed telemetry available for Schedule Form. Within that scope, review is concentrated on a much smaller set of candidate predicates and semantic annotations than the raw FSM size might suggest.

\FloatBarrier

\subsection{Failure-Signal Normalization and Snapshot-Depth Efficiency}
\label{app:signal-robustness}

This subsection checks two support claims: whether normalized failure signals preserve the same recovery decision at fixed sites, and whether the registry-only sidecar remains storage-efficient as checkpoint depth grows.

\begin{table*}[!htbp]
\caption{Failure-signal normalization adequacy matrix for the current fixed-site evaluation. At each fixed site, varying the raw observable signal leaves both the admissibility decision and the recovery signature unchanged.}
\label{tab:signal-robustness}
\centering
\scriptsize
\setlength{\tabcolsep}{3pt}
\begin{adjustbox}{max width=\textwidth}
\begin{tabular}{llllllll}
\toprule
Fixed site & Domain & Decision stable & Recovery stable & Expected decision & Checkpoint & Raw rows & Eval mode \\
\midrule
\texttt{nav\_finalize\_route\_admitted} & Navigation & yes & yes & \texttt{eligible} & \texttt{commit} & 4 & \texttt{real\_action\_error\_injection} \\
\texttt{schedule\_resolve\_slot\_admitted} & Schedule Form & yes & yes & \texttt{eligible} & \texttt{entry} & 3 & \texttt{real\_action\_error\_injection} \\
\texttt{diagnosis\_resolve\_check\_admitted} & Diagnosis & yes & yes & \texttt{eligible} & \texttt{entry} & 3 & \texttt{real\_action\_error\_injection} \\
\texttt{nav\_committed\_consumer\_blocked} & Navigation & yes & yes & \texttt{committed\_consumers\_present} & none & 3 & \texttt{normalized\_failure\_projection} \\
\texttt{irreversible\_effect\_blocked} & Synthetic effect site & yes & yes & \texttt{irreversible\_effect\_policy} & none & 3 & \texttt{normalized\_failure\_projection} \\
\bottomrule
\end{tabular}
\end{adjustbox}
\end{table*}

\begin{table}[!htbp]
\caption{Snapshot-depth efficiency summary for the schedule-form depth benchmark. The registry-only sidecar grows much more slowly than the inline payload, while restore cost stays in the same order of magnitude.}
\label{tab:depth-efficiency}
\centering
\scriptsize
\setlength{\tabcolsep}{3pt}
\begin{adjustbox}{max width=\linewidth}
\begin{tabular}{lr}
\toprule
Metric & Value \\
\midrule
Depth count & 5 \\
Maximum depth & 5 \\
Peak inline / registry payload ratio & 48.39$\times$ \\
Peak registry / inline restore ratio & 1.07$\times$ \\
Snapshot-manager growth & 23.00$\times$ \\
Registry-sidecar growth & 3.96$\times$ \\
Inline-payload growth & 20.88$\times$ \\
\bottomrule
\end{tabular}
\end{adjustbox}
\end{table}

\FloatBarrier

\subsection{Property-Wise Necessity Decomposition}
\label{app:property-necessity}

Table~\ref{tab:property-necessity} maps each conjunct of Eq.~\eqref{eq:formal-boundary} to the concrete counterexample or audit family that fails when it is removed.
Read together, the dependency-blocking witnesses and effect-policy forced-override audits show why admissibility is needed both to preserve committed downstream semantics and to avoid replay across irreversible effect boundaries.

\begin{table*}[!htbp]
\caption{Property-wise necessity decomposition for Eq.~\eqref{eq:formal-boundary}. Each row points to a concrete evidence family showing what breaks if the corresponding conjunct is removed.}
\label{tab:property-necessity}
\centering
\scriptsize
\setlength{\tabcolsep}{3pt}
\begin{adjustbox}{max width=\textwidth}
\begin{tabular}{>{\raggedright\arraybackslash}p{0.11\textwidth}>{\raggedright\arraybackslash}p{0.22\textwidth}>{\raggedright\arraybackslash}p{0.36\textwidth}>{\raggedright\arraybackslash}p{0.19\textwidth}}
\toprule
Conjunct & What fails without it & Current concrete signal & Evidence family \\
\midrule
\emph{Decidable} & failed-instance target becomes ambiguous & 270/270 observed rows align under full keys; the offline ambiguity benchmark identifies 10 same-entity multi-ordinal aliases, and weakened alias \texttt{ResolveSlot::slot[0]} still matches two candidates while the stale one erases the refined \texttt{Friday} value & failed-instance localization audit + instance-decidability counterexample \\
\emph{Closed} & unresolved branch is certified as an exit boundary & legal edges \texttt{WAITING\_POI\_SELECTION} \newline \texttt{$\rightarrow$STOP\_READY}; \newline \texttt{WAITING\_SLOT\_SELECTION} \newline \texttt{$\rightarrow$SLOT\_READY} emit unsafe \newline exit checkpoints when forced & wrong-boundary ablation \\
\emph{Separable} & rollback silently invalidates committed downstream work & disabling the guard drops committed consumers in navigation (1), schedule-form (2), and diagnosis (1) & committed-consumer blocking ablation \\
\emph{Controllable} & rollback crosses an irreversible effect boundary without a safe anchor & all 7 effect-policy-blocked events are unsafe to force, and none are observed false blocks & five-domain effect-policy blocking audits \\
\bottomrule
\end{tabular}
\end{adjustbox}
\end{table*}

\section{Proof Sketches for the Semantic Soundness Theorems}
\label{app:proofs}

\begin{lemma}[Legal edges do not imply recoverable boundaries]
\label{lem:legal-not-formal}
Let $u=(s_i,a,s_j)$ be a controller-legal edge, i.e., $(s_i,a,s_j)\in\delta$. If there exists a subtask instance $I$ associated with $u$ such that at least one of $\mathrm{Decidable}(u,I)$, $\mathrm{Closed}(u,I)$, $\mathrm{Separable}(u,I)$, or $\mathrm{Controllable}(u,I)$ does not hold, then $u$ is not a recoverable boundary for $I$ under Eq.~\eqref{eq:formal-boundary}.
\end{lemma}

\begin{theorem}[Necessity of committed-consumer blocking]
\label{thm:blocking-necessary}
Let $I_p$ be a producer instance and let $I_q$ be a committed downstream consumer of $I_p$. Any local-recovery policy that rolls back $I_p$ while leaving $I_q$ committed, without compensation, invalidation, or joint rollback of $I_q$, cannot guarantee semantic equivalence of the recovered execution. Therefore, committed-consumer blocking is necessary for sound failed-instance-local rollback.
\end{theorem}

\begin{corollary}[Soundness of dependency-aware admission under conservative dependency abstraction]
\label{cor:dependency-admission-sound}
Assume A3--A4. If local rollback of a producer instance is admitted by the runtime, then there exists no committed downstream instance that would become semantically unsupported were that producer rolled back while the downstream instance remained committed.
\end{corollary}

\begin{theorem}[Maximal admissible checkpoint selection]
\label{thm:max-admissible}
Assume the stable checkpoints of $I_f$ are totally ordered by recency under Assumption~A2. If $\mathcal{A}(f)\neq\emptyset$, then Eq.~\eqref{eq:checkpoint-select} returns a unique checkpoint
\[
c^\star(f)=\max \mathcal{A}(f)
\]
Moreover: (i) $c^\star(f)$ is admissible, i.e., $c^\star(f)\in\mathcal{A}(f)$; and (ii) for any checkpoint $c' \in \mathcal{C}(I_f)$ with $c^\star(f)\prec^{(I_f)} c'$, we have $c' \notin \mathcal{A}(f)$.
\end{theorem}

\paragraph{Proof Sketch of Lemma~\ref{lem:legal-not-formal}.}
By Eq.~\eqref{eq:formal-boundary}, $\mathrm{Recoverable}(u,I)$ is the conjunction of four obligations:
$\mathrm{Decidable}(u,I)$,
$\mathrm{Closed}(u,I)$,
$\mathrm{Separable}(u,I)$, and
$\mathrm{Controllable}(u,I)$.
Therefore, if any one of these conjuncts fails for the legal edge $u$ and instance $I$, then the conjunction itself fails, and hence $\neg \mathrm{Recoverable}(u,I)$.
Controller legality certifies only controller-level reachability and does not reintroduce any missing conjunct. Thus legality is necessary for controller execution but not sufficient for recoverability.

\paragraph{Proof Sketch of Theorem~\ref{thm:blocking-necessary}.}
Assume $I_p \rightsquigarrow I_q$ under Eq.~\eqref{eq:consumer-edge}, and that $I_q$ is already committed. By Assumption~A3, the committed state of $I_q$ semantically depends on outputs produced by $I_p$. Suppose a local-recovery policy rolls back $I_p$ while leaving $I_q$ committed and without compensating, invalidating, or jointly rolling back $I_q$. Then the committed state of $I_q$ becomes semantically unsupported after the producer rollback. This violates semantic equivalence of the recovered execution and is exactly the conflict ruled out by the \emph{NoCommittedConflict} term in Eq.~\eqref{eq:safe-recover}. Therefore any sound failed-instance-local recovery policy must block such rollback.

\paragraph{Proof Sketch of Theorem~\ref{thm:max-admissible}.}
By Assumption~A2, the checkpoint set $\mathcal{C}(I_f)$ is totally ordered by recency under $\preceq^{(I_f)}$. Since $\mathcal{A}(f)\subseteq\mathcal{C}(I_f)$ and $\mathcal{A}(f)\neq\emptyset$, the maximum element $c^\star(f)=\max\mathcal{A}(f)$ exists and is unique. By construction, $c^\star(f)\in\mathcal{A}(f)$, so $c^\star(f)$ is admissible. Now let $c' \in \mathcal{C}(I_f)$ satisfy $c^\star(f)\prec^{(I_f)} c'$. If $c'$ were also admissible, then $c' \in \mathcal{A}(f)$ and $c^\star(f)$ would fail to be the maximum element of $\mathcal{A}(f)$, a contradiction. Therefore every checkpoint strictly later than $c^\star(f)$ is inadmissible. This proves that Eq.~\eqref{eq:checkpoint-select} returns the unique latest admissible checkpoint within the failed instance.

\section{Reproducibility and Additional Experimental Details}
\label{app:repro}

\paragraph{Runtime Realization Scope.}
The recovery method is realized in \dart{} as an online runtime sidecar rather than an offline trace analyzer. The implementation centers on reviewed boundary configurations, step lifting, modular named checkpoints, producer-consumer dependency tracking, and rollback selection in the online recovery path. Appendix~\ref{app:runtime-realization} gives the concrete realization details.

\subsection{Runtime Realization Details}
\label{app:runtime-realization}

We realize failed-instance identity, recoverable-boundary review, modular checkpoints, and admissibility checks in \dart{} as a runtime sidecar attached to the normal agent loop rather than as an offline trace-analysis layer. This subsection records the concrete runtime realization of the four-layer recovery pipeline described in Section~4.

\begin{figure}[t]
  \centering
  \includegraphics[width=0.8\linewidth]{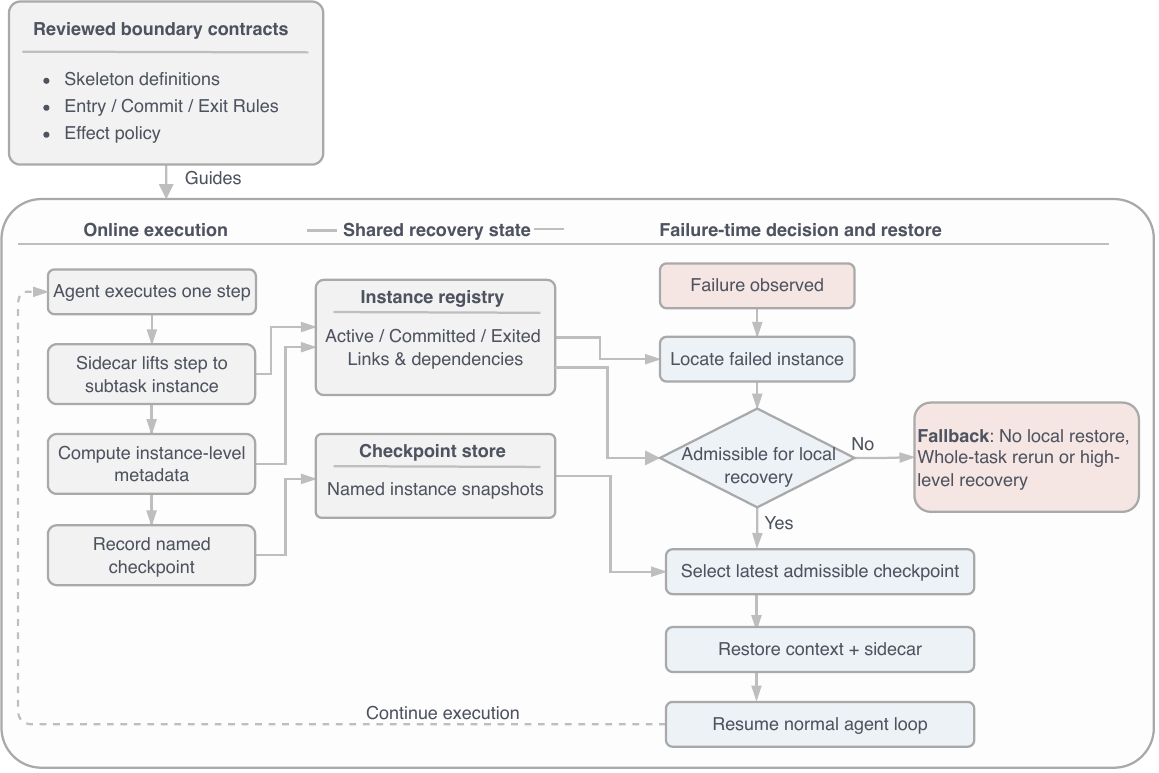}
  \caption{Runtime sidecar overview. Reviewed boundaries define recovery contracts; the sidecar lifts steps, tracks dependencies, and restores the latest admissible checkpoint.}
  \label{fig:runtime}
\end{figure}

Normalized signals use a small runtime failure vocabulary such as \texttt{TIMEOUT}, \texttt{INVALID\_OUTPUT}, or \texttt{MISSING\_INPUT}. Let $\widehat{K}(k)$ denote the reviewed configuration loaded for skeleton id $k$. In the current \dart{} realization,
\begin{equation}
\label{eq:boundary-config}
\widehat{K}(k) = \big(\widehat{S}_k^{\mathrm{int}}, \widehat{S}_k^{\mathrm{ent}}, \widehat{P}_k^{\mathrm{com}}, \widehat{P}_k^{\mathrm{exit}}, \widehat{X}_k^{\mathrm{in}}, \widehat{X}_k^{\mathrm{out}}, \widehat{\pi}_k^{\mathrm{eff}}\big)
\end{equation}
The sidecar observer then lifts a base execution step $e_t$ from Eq.~\eqref{eq:step-record} to an enriched recovery-aware record
\begin{equation}
\label{eq:step-lift}
\Psi(e_t) = (k_t, I_t, R_t, W_t, c_t)
\end{equation}
where $k_t$ is the resolved skeleton id, $I_t$ the resolved instance, $R_t$ and $W_t$ the step-level read and write sets, and $c_t$ an optional named checkpoint produced at that step. In the current system, $R_t$ and $W_t$ are assembled conservatively from explicit state/action manifests, planner maps, and reviewed tool I/O or effect annotations associated with the active skeleton, rather than inferred from arbitrary black-box runtime semantics.

At the instance level, the runtime aggregates conservative interfaces as
\begin{equation}
\label{eq:instance-rw}
R(I)=\bigcup_{t:I_t=I} R_t,
\qquad
W(I)=\bigcup_{t:I_t=I} W_t
\end{equation}
Eq.~\eqref{eq:instance-rw} is the runtime abstraction used to derive dependency edges between instances: writes summarize produced semantic objects, while reads summarize downstream consumption that may make producer rollback unsafe.
These aggregates induce the conservative producer-consumer relation
\begin{equation}
\label{eq:consumer-edge}
I_p \rightsquigarrow I_q \iff W(I_p)\cap R(I_q)\neq \varnothing
\end{equation}
Eq.~\eqref{eq:consumer-edge} realizes committed-consumer detection as a conservative over-approximation of must-block dependency. Because the loaded interfaces intentionally over-approximate possible reads and writes, the resulting dependency relation may block some otherwise admissible rollbacks; this conservatism is part of the safety envelope studied here.

In the default \registryonly{} path, the sidecar stores the instance registry, reconstructs snapshot-manager bookkeeping at restore time, and resumes normal execution from the chosen named checkpoint and sidecar state. Figure~\ref{fig:runtime} summarizes this path.

\paragraph{Benchmark Settings.}
The main text reports three integrated result tables over the three core LLM-driven domains. Semi-real live benchmarks inject failures only at observable action boundaries of the form in Eq.~\eqref{eq:failure-event}, and the headline aggregates use repeat~=~5 under the official and commitment-sensitive regimes. Appendix~\ref{app:generalization} documents the full five-domain case universe and deterministic ETL/travel generalization; Appendix~\ref{app:stats} adds paired statistics, no-failure-path overhead, and checkpoint-granularity diagnostics; Appendix~\ref{app:langgraph} provides two-domain external LangGraph evidence, including regime-aware comparison, transplant-control transportability, and the blocking witness; and Appendix~\ref{app:audit} collects the broader audit chain.

\paragraph{Representative Reproduction Path.}
An accompanying public artifact is available at \url{https://github.com/KeoYang/DART}. It includes the \dart{} implementation used in the paper, together with the reviewed boundary configurations, benchmark harnesses, and scripts used to validate the frozen paper artifacts and rerun the official non-live pipelines. The artifact exposes fixed case sets, automated interaction choices, explicit failure injection points, and scripts for regenerating the aggregate JSON and markdown result files used for the paper tables. Rerunning the semi-real live protocol further requires the corresponding hosted-LLM and, for the navigation domain, map-service credentials. For the T8 depth benchmark, the two compared variants correspond in the artifact to the aliases \registryonlyartifact{} and \inlinecounterfactualartifact{}, respectively. In particular, the artifact allows readers to trace how reviewed boundary configurations instantiate Eq.~\eqref{eq:boundary-config}, how execution steps are lifted according to Eq.~\eqref{eq:step-lift}, how committed-consumer relations are accumulated under Eq.~\eqref{eq:consumer-edge}, and how rollback targets are selected through Eq.~\eqref{eq:checkpoint-select}.

\paragraph{Compute Resources.}
The reported experiments were executed on a single local workstation with an Apple M3 CPU (8 cores) and 16~GB unified memory. The semi-real runs are lightweight in restoration cost; the dominant runtime cost comes from downstream re-execution and external service latency rather than from checkpoint restore itself.

\paragraph{Assets and Services.}
The semi-real benchmarks use hosted LLM APIs and, in the navigation domain, a map-search API. These external services are accessed only through their normal provider interfaces and terms. The public artifact does not redistribute those services or their proprietary outputs; instead, it provides the benchmark harnesses, reviewed boundary configurations, and analysis scripts needed to reproduce the reported measurements for readers with appropriate access credentials.

\paragraph{LLM Usage.}
The evaluated domains instantiate the proposed method with LLM-based tool agents under explicit FSM controllers. The contribution of the paper is not a new language model or prompting method. Instead, the proposed method operates at the runtime-recovery layer and requires only that the agent expose explicit state transitions, action boundaries, and step histories.

\end{document}